\documentclass{article}

\usepackage{microtype}
\usepackage{graphicx}
\usepackage{subcaption}
\usepackage{booktabs} % for professional tables
\usepackage[utf8]{inputenc} 
\usepackage{multirow} 
\usepackage{makecell}
\usepackage{float} 
\usepackage{hyperref}
\usepackage{tcolorbox}
\tcbuselibrary{breakable}
\usepackage{enumitem} 

\usepackage[preprint]{icml2026}

\usepackage{amsmath}
\usepackage{amssymb}
\usepackage{mathtools}
\usepackage{amsthm}
\usepackage{enumitem}
\usepackage{tabularx}
\usepackage{booktabs}
\usepackage{multirow}
\usepackage{subcaption}
\usepackage[capitalize,noabbrev]{cleveref}

\theoremstyle{plain}

\theoremstyle{definition}

\theoremstyle{remark}

\usepackage[textsize=tiny]{todonotes}
\usepackage{float} 
\usepackage{graphicx}

\icmltitlerunning{Preprint}

\begin{document}

\twocolumn[
  \icmltitle{WorldArena: A Unified Benchmark for Evaluating Perception \\ and Functional Utility of Embodied World Models}



% Equal contribution symbol
\icmlsetsymbol{equalone}{*}
\icmlsetsymbol{equaltwo}{$\ddagger$}
\icmlsetsymbol{lead}{$\S$}
\icmlsetsymbol{cor}{$\dagger$}

\begin{icmlauthorlist}
  \icmlauthor{Yu Shang}{tsinghua,equalone,lead}
  \icmlauthor{Zhuohang Li}{tsinghua,equalone}
  \icmlauthor{Yiding Ma}{tsinghua,equalone}
  \icmlauthor{Weikang Su}{tsinghua,equalone}
  \icmlauthor{Xin Jin}{equaltwo}
  \icmlauthor{Ziyou Wang}{tsinghua,equaltwo}
  \icmlauthor{Lei Jin}{tsinghua}
  \icmlauthor{Xin Zhang}{tsinghua}
  \icmlauthor{Yinzhou Tang}{tsinghua}
  \icmlauthor{Haisheng Su}{sjtu}
  \icmlauthor{Chen Gao}{tsinghua}
  \icmlauthor{Wei Wu}{tsinghua}
  \icmlauthor{Xihui Liu}{hku}
  \icmlauthor{Dhruv Shah}{princeton}
  \icmlauthor{Zhaoxiang Zhang}{cas}
  \icmlauthor{Zhibo Chen}{ustc} 
  \icmlauthor{Jun Zhu}{tsinghua}
  \icmlauthor{Yonghong Tian}{pku}
  \icmlauthor{Tat-Seng Chua}{nus}
  \icmlauthor{Wenwu Zhu}{tsinghua}
  \icmlauthor{Yong Li}{tsinghua,cor}
\end{icmlauthorlist}

\icmlaffiliation{tsinghua}{Tsinghua University, Beijing, China}
\icmlaffiliation{sjtu}{Shanghai Jiao Tong University, Shanghai, China}
\icmlaffiliation{hku}{The University of Hong Kong, Hong Kong SAR, China}
\icmlaffiliation{princeton}{Princeton University, Princeton, NJ, USA}
\icmlaffiliation{cas}{Chinese Academy of Sciences, Beijing, China}
\icmlaffiliation{ustc}{University of Science and Technology of China, Hefei, China}
\icmlaffiliation{pku}{Peking University, Beijing, China}
\icmlaffiliation{nus}{National University of Singapore, Singapore}

\icmlEqualContribution{\textsuperscript{*} Equal contribution.}
\icmlEqualContribution{\textsuperscript{$\ddagger$} Equal contribution.}
\icmlEqualContribution{\textsuperscript{$\S$} Project lead.}

\icmlcorrespondingauthor{Yong Li}{liyong07@tsinghua.edu.cn}

  \icmlkeywords{Machine Learning, ICML}

  \vskip 0.3in
]



\printAffiliationsAndNotice{}  % no special notice (required even if empty)

\begin{abstract}
While world models have emerged as a cornerstone of embodied intelligence by enabling agents to reason about environmental dynamics through action-conditioned prediction, their evaluation remains fragmented. 
Current evaluation of embodied world models has largely focused on perceptual fidelity (e.g., video generation quality), overlooking the functional utility of these models in downstream decision-making tasks. 
In this work, we introduce WorldArena, a unified benchmark designed to systematically evaluate embodied world models across both perceptual and functional dimensions. WorldArena assesses models through three dimensions: video perception quality, measured with 16 metrics across six sub-dimensions; embodied task functionality, which evaluates world models as data engines, policy evaluators, and action planners integrating with subjective human evaluation. 
Furthermore, we propose EWMScore, a holistic metric integrating multi-dimensional performance into a single interpretable index.
Through extensive experiments on 14 representative models, we reveal a significant perception–functionality gap, showing that high visual quality does not necessarily translate into strong embodied task capability.
WorldArena benchmark with the public leaderboard is released at \hyperlink{https://worldarena.ai}{https://world-arena.ai}, providing a framework for tracking progress toward truly functional world models in embodied AI. 
\end{abstract}

\section{Introduction}

\begin{figure*}[t]
    \centering
    \begin{minipage}{0.42\textwidth}
        \centering
        \includegraphics[width=\textwidth]{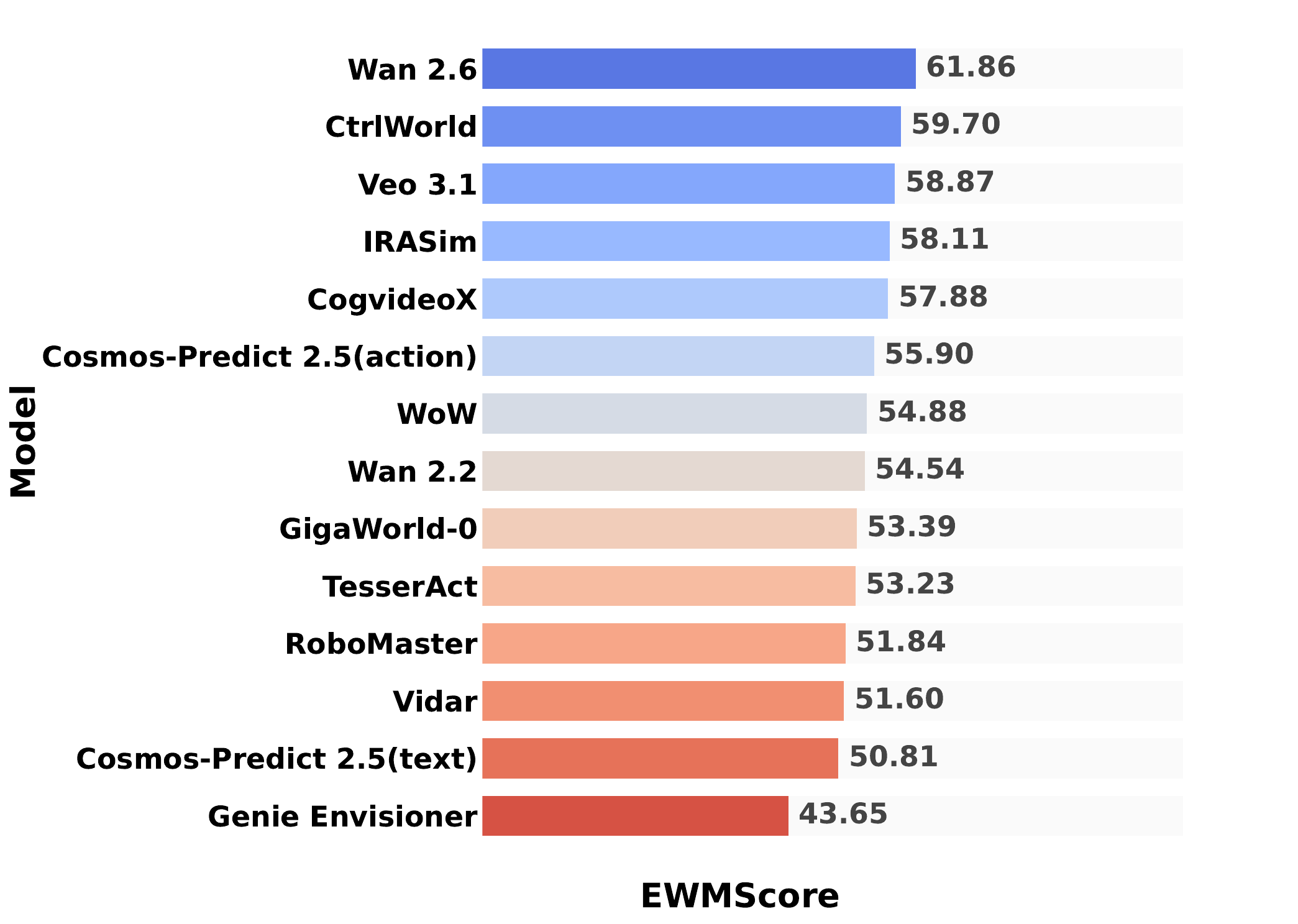}
        \subcaption{} % Label for the left image
        \label{fig:ewmscore}
    \end{minipage} \hspace{-0.05\textwidth} % Adjust the spacing between images
    \begin{minipage}{0.58\textwidth} % Adjust the width of the second image
        \centering
        \includegraphics[width=\textwidth]{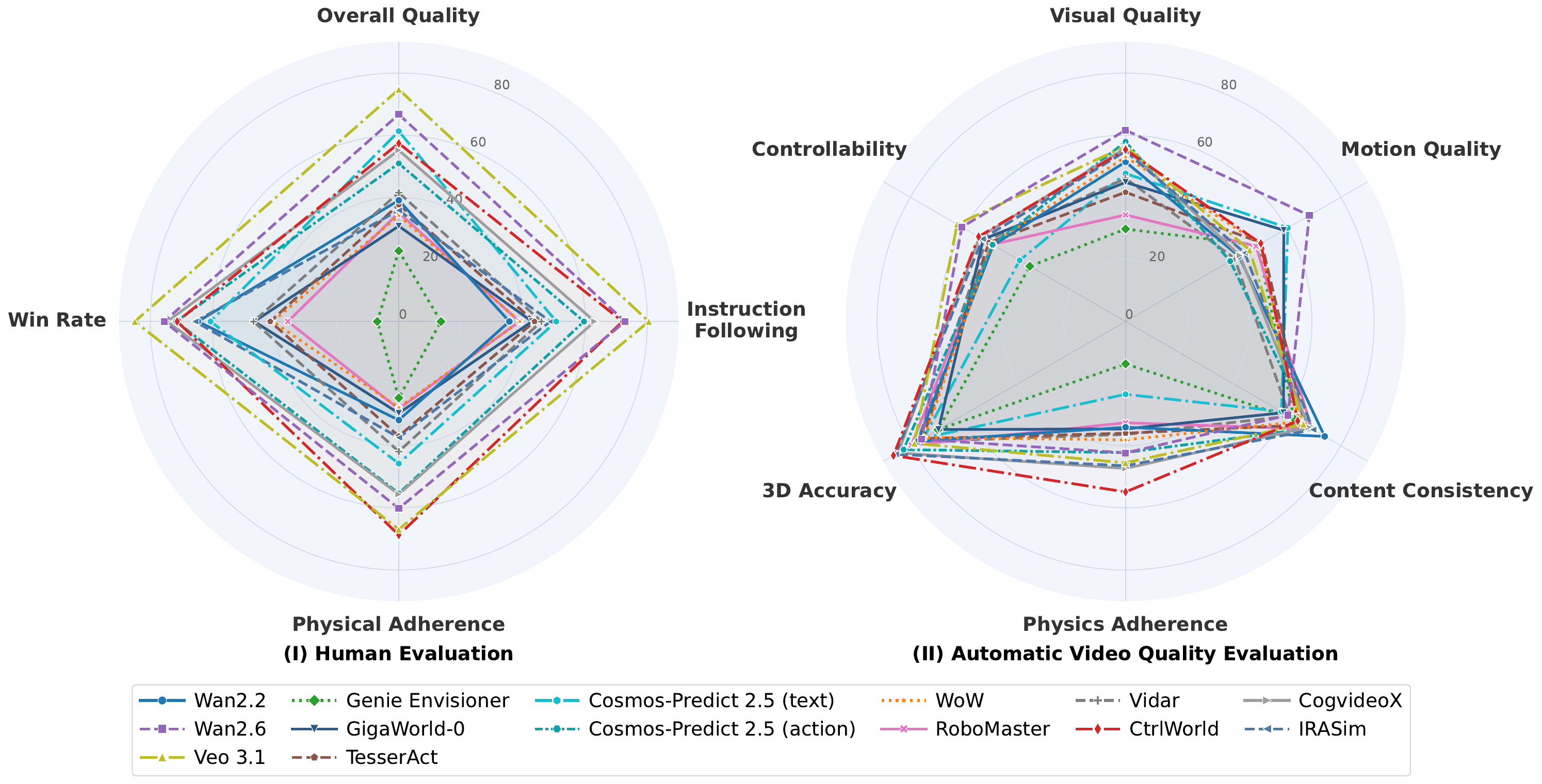}
        \subcaption{} % Label for the right image
        \label{fig:radar_chart}
    \end{minipage}
    \caption{EWMScore results (a) and performance comparisons across different evaluation dimensions (b) for 14 representative embodied world models.}

    \label{fig:combined}
\end{figure*}

In recent years, world models~\cite{ding2025understanding,kong20253d,zhu2024sora} have emerged as a foundational component of embodied intelligence.
A world model (WM) learns to predict future environment states conditioned on current observations and actions, enabling agents to reason about dynamics and interaction outcomes.
Embodied World Model (EWM) forecasts future states based on robot actions and external instructions, effectively functioning as a mental simulator guiding robot action planning and decision-making, or an environment proxy to support scalable robotic training and evaluation~\cite{shangsurvey,long2025survey}.
% Unlike general-purpose video generation models, EWMs must capture not only perceptual fidelity but also physically grounded, action-consistent dynamics essential for downstream embodied tasks.
% In embodied AI, world models can function both as a proxy for the environment, used for training and evaluating agents, and as the robot brain, guiding action planning and decision-making based on a deep understanding of environmental dynamics~\cite{shangsurvey,long2025survey}.
% Embodied world models aim to generate realistic robot interaction scenarios conditioned on external instructions.
Unlike general-purpose video generation models, EWMs must capture not only perceptual fidelity but also physically grounded, action-consistent dynamics that are critical for downstream embodied tasks. 

However, existing evaluation protocols suffer from significant limitations.
First, they lack comprehensive evaluations oriented toward embodied tasks, including the role of world models as environment proxies and embodied agents. Current benchmarks~\cite{yue2025ewmbench,li2025worldmodelbench,lu20254dworldbench} mainly focus on video-level quality metrics, which fail to reflect the real-world value of embodied world models for practical embodied applications, as shown in the comparison of Table~\ref{tab:benchmark_comparison}. Although some recent studies~\cite{qin2024worldsimbench,zhang2025world,fan2026wow} evaluate embodied world models through closed-loop action execution, broader embodied capabilities such as their roles as synthetic data engines or tools for policy evaluation remain largely unassessed.
Second, the coverage of evaluated models is insufficient. Most existing benchmarks~\cite{yue2025ewmbench,fan2026wow} focus on general text-conditioned video generation models~\cite{wan2025wan,yang2024cogvideox}, while many recent robot-specialized world models~\cite{chi2025wow,team2025gigaworld,liao2025genie,zhen2025tesseract,guo2025ctrl}, have received little systematic attention and remain largely unevaluated.

To bridge this gap, we present \textbf{WorldArena}, the first embodied world model benchmark that integrates perceptual and three functional evaluations, combining both objective and subjective assessments.
WorldArena provides a holistic evaluation framework across three complementary aspects:
(1) multi-faceted video quality, comprising 16 numerical metrics across 6 key sub-dimensions, including visual quality, motion quality, content consistency, physics adherence, 3D accuracy, and controllability;
(2) embodied task utility, which evaluates model performance in data synthesis, policy evaluation, and action planning; and
(3) human evaluation, which complements automated metrics by capturing qualitative aspects of model behavior that are difficult to quantify, such as physical plausibility and instruction adherence.
Additionally, we introduce EWMScore, a unified metric that combines multi-dimensional metrics into a single index, offering a comprehensive assessment of embodied world models' generative performance.
An overview of the evaluation result is shown in Figure~\ref{fig:combined}.

For the evaluation data, we select bimanual robotic manipulation as a representative embodied scenario and conduct evaluations based on the RobotTwin 2.0 dataset~\cite{chen2025robotwin}, which covers 50 diverse robotic scenarios, ensuring both scenario diversity and evaluation reliability. 
We perform a unified evaluation on 14 representative world models, including both general video generation world models and specialized embodied world models. 
The results reveal a significant gap between visual fidelity and embodied task performance, indicating that current visual quality has not yet reached the level required to effectively support embodied tasks.
Overall, our main contributions can be summarized as follows:
\begin{itemize}[leftmargin=*]
    \item We introduce the first comprehensive benchmark tailored for embodied world models, enabling a unified evaluation of their perceptual and functional capabilities.
    \item We propose EWMScore, a unified objective metric for embodied world models, and conduct extensive human studies to validate its effectiveness. Results demonstrate that EWMScore highly aligns with subjective judgment, serving as a reliable and interpretable index.
    \item We conduct a systematic evaluation of 14 representative embodied world models and provide a multi-dimensional analysis of their strengths and limitations, offering insights and guidance for future research.
\end{itemize}

\section{Related Works}

\begin{table*}[t]
\centering
\caption{Comparison of existing world model benchmarks and WorldArena across three key evaluation dimensions.}
\label{tab:benchmark_comparison}
\small
\setlength{\tabcolsep}{4pt}
\renewcommand{\arraystretch}{1.15}

\begin{tabularx}{\textwidth}{c|*{6}{>{\centering\arraybackslash}X}|*{3}{>{\centering\arraybackslash}X}|>{\centering\arraybackslash}X}
\toprule
\multirow{2}{*}{\textbf{Benchmark}} 
& \multicolumn{6}{c|}{\textbf{Video Quality}} 
& \multicolumn{3}{c|}{\textbf{Embodied Tasks}} 
& \multirow{2}{*}{\textbf{Human}} \\  
\cmidrule(lr){2-7} \cmidrule(lr){8-10}
& Visual 
& Motion 
& Content 
& Physics 
& Control 
& 3D 
& Data 
& Policy 
& Action \\
& Quality 
& Quality 
& Consist. 
& Adher. 
& ability 
& Acc. 
& Engine 
& Eval. 
& Planner \\
\midrule
WorldModelBench~\cite{li2025worldmodelbench}     & $\times$ & $\times$ & $\times$ & $\checkmark$ & $\checkmark$ & $\times$ & $\times$ & $\times$ & $\times$ & $\checkmark$ \\
WorldSimBench~\cite{qin2024worldsimbench}       & $\checkmark$ & $\checkmark$ & $\checkmark$ & $\times$ & $\checkmark$ & $\times$ & $\times$ & $\times$ & $\checkmark$ & $\checkmark$ \\
WorldScore~\cite{duan2025worldscore}          & $\checkmark$ & $\checkmark$ & $\checkmark$ & $\times$ & $\checkmark$ & $\checkmark$ & $\times$ & $\times$ & $\times$ & $\checkmark$ \\
4DWorldBench~\cite{lu20254dworldbench}        & $\checkmark$ & $\checkmark$ & $\checkmark$ & $\checkmark$ & $\checkmark$ & $\checkmark$ & $\times$ & $\times$ & $\times$ & $\checkmark$ \\
EWMBench~\cite{yue2025ewmbench}            & $\times$ & $\checkmark$ & $\checkmark$ & $\times$ & $\checkmark$ & $\times$ & $\times$ & $\times$ & $\times$ & $\checkmark$ \\
WorldEval~\cite{li2025worldeval}           & $\times$ & $\times$ & $\times$ & $\times$ & $\times$ & $\times$ & $\times$ & $\checkmark$ & $\times$ & $\times$ \\
World-in-World~\cite{zhang2025world}      & $\checkmark$ & $\checkmark$ & $\times$ & $\times$ & $\checkmark$ & $\times$ & $\times$ & $\times$ & $\checkmark$ & $\times$ \\
WoW-World-Eval~\cite{fan2026wow}      & $\checkmark$ & $\checkmark$ & $\checkmark$ & $\checkmark$ & $\checkmark$ & $\times$ & $\times$ & $\times$ & $\checkmark$ & $\checkmark$ \\
\midrule
\textbf{WorldArena (Ours)} 
& $\checkmark$ & $\checkmark$ & $\checkmark$ & $\checkmark$ & $\checkmark$ & $\checkmark$ 
& $\checkmark$ & $\checkmark$ & $\checkmark$ 
& $\checkmark$ \\
\bottomrule
\end{tabularx}
\end{table*}

\subsection{Embodied World Models}
Embodied world models are generative models that predict future observations of physical scenes involving robot locomotion and manipulation. These models can be broadly categorized into three types: video generation-based models~\cite{liao2025genie,shang2025roboscape,team2025gigaworld}, 3D reconstruction-based models~\cite{huang2025enerverse,qian2025wristworld}, and latent-space world models~\cite{assran2025v,liu2025jepa,hafner2025mastering}.
In practice, embodied world models serve three key roles: (1) as data synthesis engines~\cite{jang2025dreamgen} for generating video-action sequences to augment robot policy training; (2) as policy evaluation environments~\cite{shang2025roboscape,li2025worldeval} for scalable virtual testing through policy-world model interaction; and (3) as action planners~\cite{hu2024video,fan2026wow}, where predicted states are decoded into executable actions for robot control.
Given the diversity of model paradigms and functional roles, embodied world models are inherently challenging to evaluate comprehensively, underscoring the need for a holistic benchmark that systematically assesses them across both perceptual and functional dimensions, driving their future development.

\subsection{World Model Benchmarks}
Existing benchmarks for world models can be broadly categorized into general-purpose and embodied benchmarks. General-purpose benchmarks~\cite{duan2025worldscore,li2025worldmodelbench,lu20254dworldbench} primarily evaluate world models from a perceptual and generative perspective, focusing on video quality aspects such as visual fidelity, motion realism, content consistency, and, in some cases, physical plausibility or geometric consistency. While these benchmarks are effective for standardizing generative evaluation, they largely treat world models as video generators and do not assess their functional roles in decision-making or interaction. More recent embodied world model benchmarks~\cite{yue2025ewmbench,li2025worldeval,zhang2025world,fan2026wow} extend evaluation to controllability, action conditioning, and limited closed-loop interaction. However, existing embodied benchmarks remain limited in scope, often focusing on a single embodied role and predominantly targeting text-conditioned video models, with insufficient coverage of action-conditioned and robot-centric world models. 
Moreover, most existing benchmarks evaluate fewer than ten models, which further limits the scope and comprehensiveness.
In contrast, WorldArena provides a unified benchmark that systematically evaluates embodied world models across both perceptual and functional dimensions, integrating objective metrics with human subjective assessments.

\section{The WorldArena Benchmark}
The evaluation framework of WorldArena consists of three key components. First, we assess video quality from 6 dimensions with 16 metrics, focusing on the world model’s open-loop prediction ability (Section~\ref{sec:3_1}). Second, we evaluate the world model’s closed-loop performance across 3 typical embodied downstream tasks (Section~\ref{sec:3_2}). Third, to complement objective measurements with subjective judgment, we collect human annotations to assess qualitative aspects of model performance (Section~\ref{sec:3_3}).
Finally, we integrate multi-dimensional video metrics into an interpretable index EWMScore to reflect overall performance (Section~\ref{sec:3_4}).

\subsection{Video Quality Evaluation} \label{sec:3_1}
We begin by evaluating the quality of the videos generated by different embodied world models, considering 16 video metrics across six sub-dimensions, as shown in Figure~\ref{fig:combined} (b).
The detailed metric explanations can be found in Appendix~\ref{appendix:metric} and the case visualization is shown in Appendix~\ref{appendix:metric visualization}.

\subsubsection{Visual Quality}
Visual quality assesses whether generated videos are perceptually reliable for embodied scenarios, considering low-level fidelity, perceptual appeal, and similarity to real data. We evaluate it using three metrics:

\textbf{Image Quality} measures clarity and sharpness of frames using the MUSIQ~\cite{ke2021musiq} model, which detects distortions such as overexposure, noise, and compression artifacts~\cite{huang2024vbench}. Higher scores indicate cleaner and more coherent images.
% We present an example in Figure \ref{fig:Vision Quality} in Appendix \ref{appendix:metric visualization} and leave more details in Appendix \ref{appendix:image quality}.

\textbf{Aesthetic Quality} evaluates the visual appeal of the video, considering lighting and color composition. Using the LAION aesthetic predictor~\cite{laion2022aesthetic}, we map frames to an aesthetic feature space and derive an average score~\cite{huang2024vbench}, capturing both perceptual consistency and artistic quality.
% We present an example in Figure \ref{fig:Vision Quality} in Appendix \ref{appendix:metric visualization} and leave more details in Appendix \ref{appendix:aesthetic quality}.

\textbf{JEPA Similarity} quantifies similarity between feature distributions extracted by the pretrained V-JEPA encoder~\cite{bardes2023v}, using maximum mean discrepancy (MMD) with a second-order polynomial kernel~\cite{luo2024jedi}. Higher values indicate greater similarity to the ground-truth video.

\subsubsection{Motion Quality}
Motion quality reflects whether a model captures physically meaningful and temporally coherent dynamics. We assess both the strength of motion and its temporal continuity. To this end, we introduce the following three metrics:

\textbf{Dynamic Degree} quantifies the motion intensity within the video. Using the RAFT~\cite{teed2020raft} optical flow model, we extract motion vector fields between consecutive frames and focus on the top 5\% of active pixels~\cite{huang2024vbench}. A higher dynamic degree score indicates more pronounced and meaningful movement in the video, capturing the intensity of motion in key areas such as robotic arm gestures.
% We present an example in Figure \ref{fig:Motion Quality} in Appendix \ref{appendix:metric visualization} and leave more details in Appendix \ref{subsec:dynamic}.

\textbf{Flow Score} measures the overall intensity of motion across the video by averaging optical flow magnitudes over time~\cite{liu2023evalcrafter}. This score reflects the degree of dynamic interaction, where a higher value indicates greater motion intensity and more physically meaningful dynamics throughout the video.

\textbf{Motion Smoothness} evaluates the temporal coherence of motion, assessing whether movements between consecutive frames are smooth and consistent with physical inertia. Using a frame interpolation model~\cite{zhang2024vfimambavideoframeinterpolation}, we predict intermediate frames and compare them to real frames~\cite{duan2025worldscore}. This approach incorporates motion magnitude as a weighting factor to prevent overestimating static backgrounds and ensure rapid motion sequences are not unfairly penalized.
% We present an example in Figure \ref{fig:Motion Quality} in Appendix \ref{appendix:metric visualization} and leave more details in Appendix \ref{appendix:flow score}.

% \textbf{Motion Smoothness} evaluates the temporal coherence of motion, judging whether movements between consecutive frames are smooth and consistent with physical inertia. We use a frame interpolation model~\cite{zhang2024vfimambavideoframeinterpolation} to predict intermediate frames and compare them to the real intermediate frames~\cite{duan2025worldscore}. This approach incorporates motion magnitude as a weighting factor, helping avoid overestimating static backgrounds and ensuring that rapid motion sequences are not penalized unfairly.
% We present an example in Figure \ref{fig:Motion Quality} in Appendix \ref{appendix:metric visualization} and leave more details in Appendix \ref{appendix:motion smoothness}.

\subsubsection{Content Consistency}
Content consistency measures the stability of objects and scenes throughout the video, evaluated at both semantic and appearance levels using three metrics:

\textbf{Subject Consistency} assesses object consistency across frames by calculating cosine similarity between DINO~\cite{caron2021emerging} features from the first, current, and previous frames~\cite{huang2024vbench}. Higher similarity scores indicate better consistency.
% We present an example in Figure \ref{fig:Content Consistency} in Appendix \ref{appendix:metric visualization} and leave more details in Appendix \ref{appendix:subject consistency}.

\textbf{Background Consistency} evaluates scene stability using CLIP~\cite{radford2021learningtransferablevisualmodels} features, measuring cosine similarity between the current frame and the first and previous frames to assess scene stability~\cite{huang2024vbench}.
% We present an example in Figure \ref{fig:Content Consistency} in Appendix \ref{appendix:metric visualization} and leave more details in Appendix \ref{appendix:background consisitency}.

\textbf{Photometric Consistency} measures pixel-level texture stability by calculating the average end-point  error (AEPE) using optical flow~\cite{duan2025worldscore}. A higher AEPE indicates poorer alignment, while a higher score reflects better consistency.

\subsubsection{Physics Adherence}
Physics adherence evaluates whether generated behaviors conform to real-world physical constraints rather than merely appearing visually plausible.
We therefore assess both local interaction realism and global motion correctness with the following two metrics:

\textbf{Interaction Quality} evaluates the physical plausibility of interactions between the robot and objects. We use Qwen3-VL~\cite{Qwen3-VL} to assess factors such as contact behavior and force transmission, checking whether the interactions are physically realistic. The interaction quality score is based on a 1–5 scale, normalized to [0,1], showing how well the robot’s actions align with expected physical behaviors.

\textbf{Trajectory Accuracy} quantifies the accuracy of the robotic arm's grasping trajectory. Using the SAM 3~\cite{carion2025sam} model, we extract bounding boxes for the arm in each frame and compute the normalized dynamic time warping (NDTW) distance to evaluate alignment with the ground-truth trajectory~\cite{yue2025ewmbench}. A higher score reflects better spatial-temporal alignment and more accurate trajectory prediction.
% We present an example in Figure \ref{fig:Physics Adherence} in Appendix \ref{appendix:metric visualization} and leave more details in Appendix \ref{appendix:interaction quality}.

% \textbf{Trajectory Accuracy} quantifies the accuracy of the robotic arm's grasping trajectory. Using the SAM 3~\cite{carion2025sam} model, we extract bounding boxes for the robotic arm in each frame and construct the trajectory sequence. We then compute the normalized dynamic time warping (NDTW) distance~\cite{book} to evaluate the alignment between the generated and ground-truth trajectories~\cite{yue2025ewmbench}. A higher score indicates better spatial-temporal alignment with the true trajectory, reflecting more accurate trajectory prediction.
% We present an example in Figure \ref{fig:Physics Adherence} in Appendix \ref{appendix:metric visualization} and leave more details in Appendix \ref{appendix:trajectory quality}.

\subsubsection{3D Accuracy}
3D accuracy assesses whether generated videos preserve real-world spatial structure beyond image appearance. We evaluate geometric consistency and perspective plausibility with the following two metrics:

\textbf{Depth Accuracy} evaluates whether the generated video preserves real-world spatial geometry by comparing depth maps between the generated and ground-truth videos. We use monocular depth estimation and apply a median-based scaling strategy to address scale ambiguity. A higher depth accuracy score indicates better geometric consistency with the real-world scene.
% We present an example in Figure \ref{fig:3D Accuracy} in Appendix \ref{appendix:metric visualization} and leave more details in Appendix \ref{appendix:depth quality}.

\textbf{Perspectivity} evaluates the 3D plausibility of the video, focusing on factors such as scale variation with depth, lighting consistency, and occlusion relationships. We use Qwen3-VL as a judge to assess the perspective, judging whether the video adheres to realistic 3D geometry. A higher score reflects better perspective alignment with real-world scenes.
% We present an example in Figure \ref{fig:3D Accuracy} in Appendix \ref{appendix:metric visualization} and leave more details in Appendix \ref{appendix:perspectivity}.

\subsubsection{Controllability}
Controllability measures the model's ability to respond to external instructions. We evaluate whether generated videos align with intended actions and instructions using three metrics:

\textbf{Instruction Following} assesses the model's accuracy in following instructions regarding action type, target object, and task state, measured by a VLM-based judge (Qwen3-VL) and scores normalized to [0,1].
% We present an example in Figure \ref{fig:Controllability} in Appendix \ref{appendix:metric visualization} and leave more details in Appendix \ref{appendix:instruction following}.

\textbf{Semantic Alignment} measures how well the generated video matches the semantic meaning of the instruction by computing cosine similarity between Qwen2.5-VL-generated~\cite{Qwen2.5-VL} descriptions of the generated and reference videos.
% We present an example in Figure \ref{fig:Controllability} in Appendix \ref{appendix:metric visualization} and leave more details in Appendix \ref{appendix:semantic alignment}.

\textbf{Action Following} evaluates video diversity in response to different instructions. For a given initial frame, we automatically generate three distinct instructions and then use the world model to generate corresponding videos.  
The diversity score is the average pairwise feature dissimilarity, with higher values indicating greater diversity.

\begin{figure}[t]
    \centering
    \includegraphics[width=0.5\textwidth]{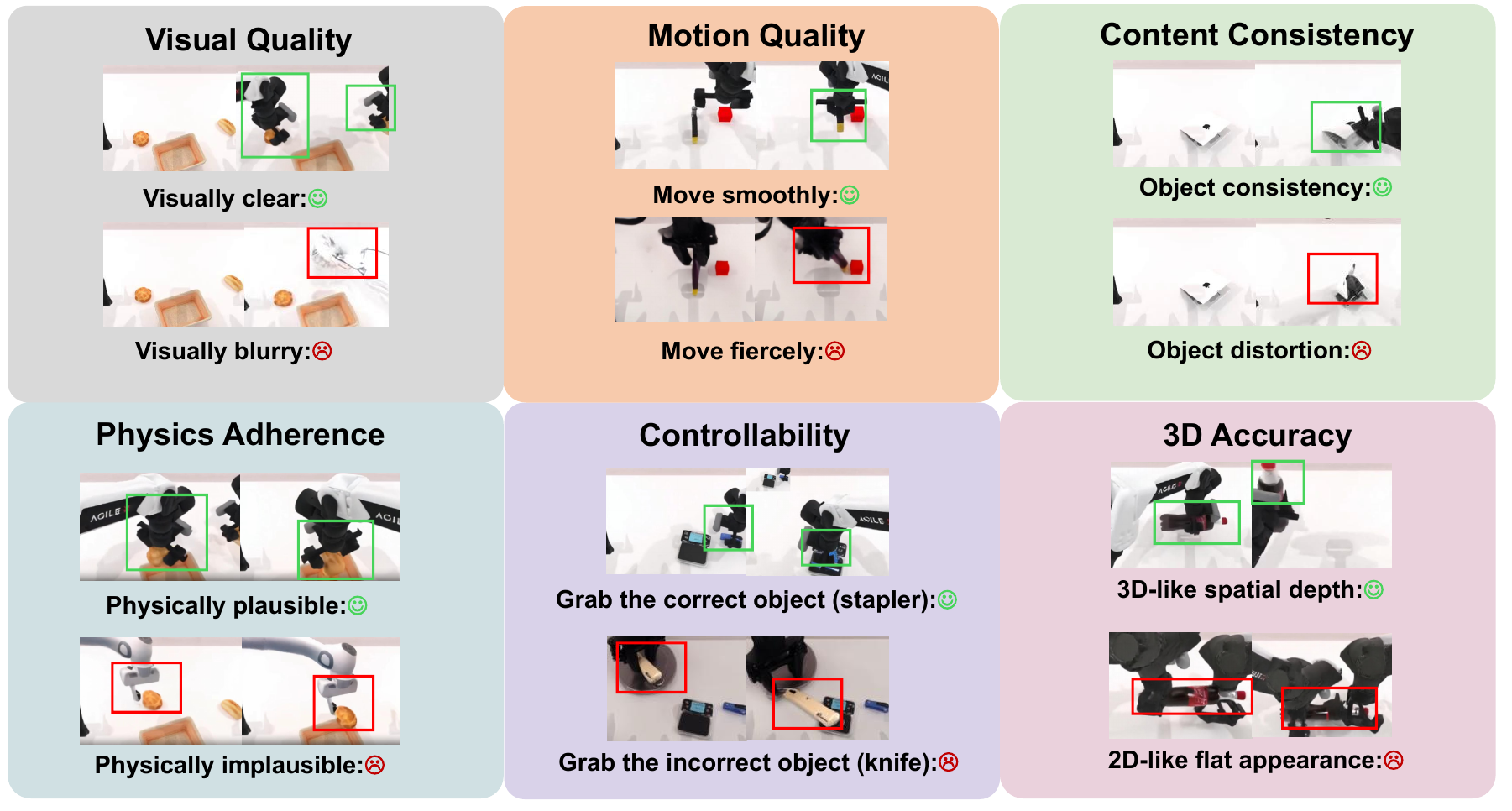}
    \caption{Illustrations of the video quality evaluations across six dimensions: visual quality, motion quality, content consistency, physics adherence, 3D accuracy, and controllability.}
    \label{fig:embtask}
\end{figure}

\subsection{Embodied Task Evaluation} \label{sec:3_2}
In this section, we evaluate the capabilities of world models through three embodied tasks, as illustrated in Figure~\ref{fig:embtask}.

\textbf{Embodied Data Engine.}
World models can generate future observations based on external instructions, enabling synthetic data generation to supplement training data for downstream embodied policy models and alleviate the scarcity of real-world data. In this part, we treat world models as embodied data synthesis engines and evaluate their performance by measuring the gain they provide to policy models.
We employ a two-phase training procedure. In the first phase, we fine-tune the world model on the RobotTwin 2.0 dataset and generate synthetic videos conditioned on the first frame and external instructions. In the second phase, we freeze the world model’s weights and integrate an inverse dynamics model (IDM) to extract actions from video features. Specifically, we follow the VPP~\cite{hu2024video} design of the diffusion policy head, guiding an action denoising head with intermediate world model features for action prediction. This process produces paired video-action sequences.
We then evaluate the impact of world model–generated synthetic data by training a baseline $\pi_{0.5}$~\cite{intelligence2025pi05visionlanguageactionmodelopenworld} policy model with varying amounts of synthetic data. The performance gain of the policy model reflects the world model’s capability to enhance policy learning.

\textbf{Embodied Policy Evaluator.}
In this section, we assess the capability of world models as environment proxies for evaluating policy performance. We train a series of policy models ($\pi_{0.5}$) with varying capabilities using the RoboTwin 2.0 dataset. These models are evaluated by interacting with an action-controllable world model, generating observation videos through a rollout process that continues until it exceeds 20\% more frames than the corresponding ground truth video. Task success is evaluated using a VLM, which determines whether the embodied task was executed successfully. The used prompt for the VLM is shown in Appendix~\ref{appendix:policy_judge_prompt}.
The success rate from the world model's evaluation is compared to that from the RoboTwin simulator. A high correlation between the two suggests effective simulation of real-world dynamics, while a low correlation indicates a mismatch in environmental transition simulation.

\textbf{Embodied Action Planner.}
By predicting future state transitions, world models can function as the action-planning "brain" of an embodied agent. In this part, we investigate the ability of world models to execute embodied tasks in a closed-loop manner. Similar to the data synthesis engine setup, we pair the world model with an inverse dynamics model, where the world model takes textual instructions and the initial frame as input and outputs the corresponding action sequence for future operations. This sequence is then executed in the RoboTwin simulator, and the task success rate is measured to evaluate the world model's performance in closed-loop action execution.

\begin{figure*}[t]
    \centering
    \includegraphics[width=0.9\textwidth]{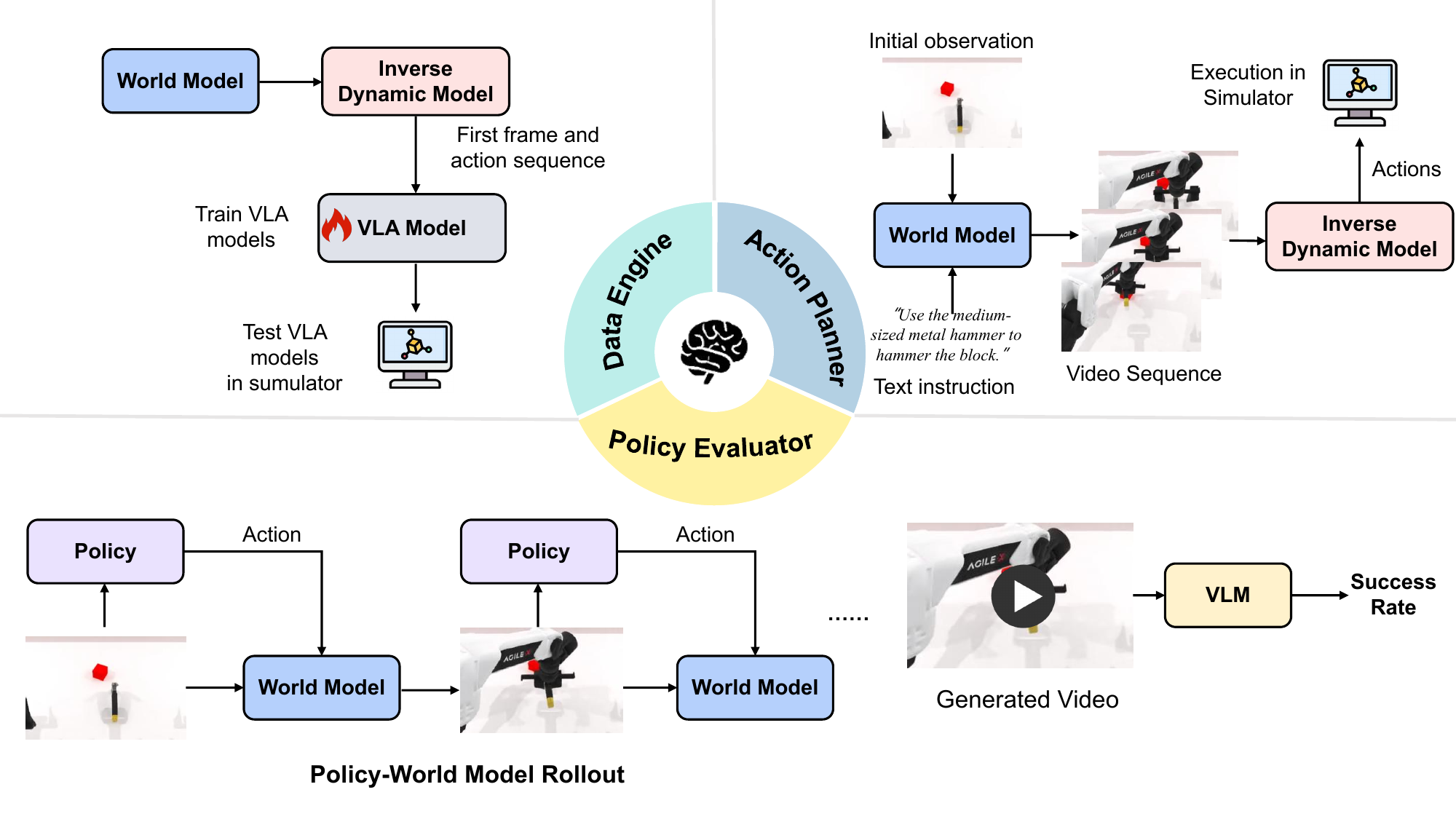}
    \caption{Overview of the embodied task evaluation systems, including the assessment of world models as embodied data engines (measuring success rate of trained downstream policies), policy evaluators (measuring correlation between world model and real-world evaluation results), and action planners (measuring success rate of world model-based policies).}
    \label{fig:embtask}
\end{figure*}

\subsection{Human Evaluation} \label{sec:3_3}
Since video quality metrics alone cannot fully capture aspects like physical plausibility and instruction adherence, we incorporate two types of human evaluations. The first type involves scoring three key dimensions: overall video quality, instruction following, and physical adherence on a 1 to 5 scale, then normalizing to a 0-100 range. The second type is a head-to-head comparison, where annotators choose the superior video generated by two different models from the same prompt, yielding a win-rate metric. We recruited 70 annotators who evaluated a total of 3500 videos.

\subsection{EWMScore Metric} \label{sec:3_4}
After computing the 16 video quality metrics spanning six perceptual dimensions, we apply a linear normalization based on empirically defined metric boundaries to map all scores into the range, and subsequently scale them to [0,100]. We then compute the arithmetic mean across all normalized metrics to obtain a single composite score, referred to as \textbf{EWMScore}.
EWMScore serves as an objective and automated metric for assessing the overall generative quality of embodied world models.

\section{Experiments}

\subsection{Experimental Setup}
\textbf{Dataset.}
We focus on robotic manipulation scenarios, using the RoboTwin 2.0~\cite{chen2025robotwin} dataset and simulator for evaluation, which includes 50 task scenarios and 2500 videos. For video quality evaluation, we use 2000 videos to train the world model and 500 videos for testing. For the embodied data engine task, we train the $\pi_{0.5}$ policy model with 10\%, 20\%, 30\%, 50\%, and 100\% of the data, resulting in a series of policy models with varying performance. For policy evaluation and action planning tasks, we conduct evaluations within the RoboTwin simulator environment.

\textbf{Tested Models.}
We evaluate 14 representative world models, covering both general-purpose video world models and embodied-specific models.
The evaluated general video world models include CogvideoX~\cite{yang2024cogvideox}, Wan 2.2~\cite{wan2025wan}, Wan 2.6~\cite{wan2025wan}, and Veo 3.1\footnote{https://aistudio.google.com/models/veo-3}.
The text-conditioned embodied world models consist of Genie Envisioner~\cite{liao2025genie}, GigaWorld~\cite{team2025gigaworld}, TesserAct~\cite{zhen2025tesseract}, Cosmos-Predict 2.5~\cite{gu2025cosmos}, WOW~\cite{chi2025wow}, RoboMaster\footnote{https://huggingface.co/datasets/robomaster2025/RoboMaster}
, Cosmos-Predict 2.5 (text)~\cite{gu2025cosmos}, and Vidar~\cite{feng2025vidar}.
In addition, we include action-conditioned embodied world models, namely IRASim~\cite{zhu2024irasim}, Cosmos-Predict 2.5 (action)~\cite{gu2025cosmos} and CtrlWorld~\cite{guo2025ctrl}.
For fair comparison, all models with available training code are post-trained on the used dataset following their official implementations.

\begin{table*}[t]
\centering
\caption{Video quality evaluation results across visual quality, motion quality and content consistency dimensions.}
\label{tab:benchmark_part1}
\small
\setlength{\tabcolsep}{6pt}
\resizebox{\textwidth}{!}{
\begin{tabular}{l|ccc|ccc|ccc}
\toprule
\multirow{2}{*}{\textbf{Models}} & \multicolumn{3}{c|}{\textbf{Visual Quality}} & \multicolumn{3}{c|}{\textbf{Motion Quality}} & \multicolumn{3}{c}{\textbf{Content Consistency}} \\
\cmidrule(lr){2-4} \cmidrule(lr){5-7} \cmidrule(lr){8-10}
 & \makecell{Image\\Quality} & \makecell{Aesthetic\\Quality} & \makecell{JEPA\\Similarity} & \makecell{Dynamic\\Degree} & \makecell{Flow\\Score} & \makecell{Motion \\ Smoothness} & \makecell{Subject \\ Consist.} & \makecell{Background \\ Consist.} & \makecell{Photometric \\ Consist.}  \\
\midrule
GigaWorld-0 &0.5041  &0.3991  &0.4413  &0.6709  &0.3118  &0.7811  &0.7303  &0.8563  &0.1756  \\
Genie Envisioner &0.2305  & 0.3289 &0.3340  &0.6930  &0.0855  &0.6966  &0.7760  &0.9024  &0.2006  \\
TesserAct&0.3322  &0.4590  &0.4579  &0.5150  &0.2447  &0.7579  &0.8250  &\textbf{0.9238}  &0.2491  \\
RoboMaster &0.3487  & 0.3842 & 0.2966 &0.6124  & 0.1484 &0.6940  & 0.8295 &0.9123  &0.3356  \\
Vidar &0.4145  &0.4068  &0.5608  & 0.2767 &0.1426  & 0.7973 & 0.7629 &0.8300  &0.2350  \\
Cosmos-Predict 2.5 (text) &0.6668  &0.4501  &0.3126  &0.5911  &0.4302  &0.7882  &0.7488  &0.8511  &0.1383  \\
Cosmos-Predict 2.5 (action) &0.4489  &0.3576  & 0.9296 &0.3994  &0.0573  &0.7100  &0.8197  &0.8894  & 0.3528 \\
WoW &0.4587  &0.3868  &0.7440  &0.4608  &0.2706  &0.7692  &0.8161  &0.9025  &0.2170  \\
CtrlWorld &0.3522  &0.3893  &0.9185  &0.4257  &0.3449  &0.7377  &\textbf{0.8411}  &0.9057  &0.1729  \\
Wan 2.2 &0.3884  &0.3963  &0.7575  & 0.4349 &0.1269  &0.7019  &0.8388  &0.9042  &\textbf{0.4776}  \\
CogvideoX &0.3582  &0.3777  & \textbf{0.9384} &0.3166  &0.2189  &0.7391  &0.8083  & 0.8773 &0.3580  \\
% WorldScape &  &  &  &  &  &  &  &  &  \\
% Hunyuanvideo 1.5 &  &  &  &  &  &  &  &  &  \\
% RoboScape &  &  &  &  &  &  &  &  &  \\
IRASim &0.3489  &0.3623 &0.9330  &0.4139  &0.2083 &0.7052 & 0.8312 &0.9068 &0.3522\\
% ivideoGPT &  &  &  &  &  &  &  &  &  \\
Veo 3.1 & 0.6605 &\textbf{0.4632}  &0.5694  &0.5450  & 0.1396 &0.6989  &0.7878  &0.8710  &0.3247  \\
Wan 2.6 &\textbf{0.6824}  &0.4433  &0.7229  &\textbf{0.7421}  &\textbf{0.4532}&\textbf{0.8539}  & 0.7517 &0.8687  &0.1904    \\
\bottomrule
\end{tabular}
}
\end{table*}

\begin{table*}[t]
\centering
\caption{Video quality evaluation results across physics adherence, 3D accuracy and controllability dimensions.}
\label{tab:benchmark_part2}
\small
\setlength{\tabcolsep}{6pt}
\begin{tabular}{l|cc|cc|cccc}
\toprule
\multirow{2}{*}{\textbf{Models}} & \multicolumn{2}{c|}{\textbf{Physics Adherence}} & \multicolumn{2}{c|}{\textbf{3D Accuracy}} & \multicolumn{3}{c}{\textbf{Controllability}} \\
\cmidrule(lr){2-3} \cmidrule(lr){4-5} \cmidrule(lr){6-8}
 & \makecell{Interaction \\ Quality} & \makecell{Trajectory \\ Acc.}  & \makecell{Depth \\ Acc.} & Perspectivity & \makecell{Instruction \\ Following} & \makecell{Semantic \\ Alignment} &  \makecell{Action \\ Following} \\
\midrule
GigaWorld-0 &0.5368  &0.1552 &0.6316  &0.7596  &0.6156 &0.8591  &0.1134  \\
Genie Envisioner &0.2052 &0.0679 &0.8663  &0.5284  &0.2028 &0.8544  &0.0109  \\
TesserAct&0.5800  &0.1396 &0.7159  &0.7920  &0.6152 &0.8783  &0.0311  \\
RoboMaster &0.5364  &0.1158 &0.8335  &0.7588  &0.5772 &0.8761  &0.0352  \\
Vidar &0.5348  &0.1928 &0.7872  &0.7592  &0.5912 &0.8826  &0.0819  \\
Cosmos-Predict 2.5 (text) &0.3872  &0.0816 & 0.7051 &0.7964  &0.2664 & 0.7733 &\textbf{0.1418}  \\
Cosmos-Predict 2.5(action) &0.5500  &0.2945 &0.8862  &0.7644  &0.5840 &0.8879  &0.0133  \\
WoW &0.5564  &0.2058 &0.7283  &0.7672  &0.5692 &0.8842  & 0.0434 \\
CtrlWorld &0.6212  &\textbf{0.4766} &0.9300  &0.7960  &0.7272 & 0.8912 &0.0210  \\
Wan 2.2 &0.5184  &0.1627 &0.7768  &0.7660  &0.5376 &0.8877  & 0.0512 \\
CogvideoX &0.5940  &0.3526 & 0.9097 &0.7828  &0.7268 &\textbf{0.8977}  &0.0076  \\
% WorldScape &  & &  &  & &  &  \\
% Hunyuanvideo 1.5 &  & &  &  & &  &  \\
% RoboScape &  & &  &  & &  &  \\
IRASim & 0.5656 &0.3639 &\textbf{0.9312}  &0.7788 &0.6604 &0.8849&0.0526  \\
% ivideoGPT &  & &  &  & &  &  \\
% Cosmos-Predict 2.5(action) &0.5500  &0.2945 &0.8862  &0.7644  &0.5840 &0.8879  &  \\
Veo 3.1 &\textbf{0.7872}  &0.1231 &0.7421  &\textbf{0.8276}  &\textbf{0.9328} & 0.8607 &0.0852  \\
Wan 2.6 & 0.7280 &0.1182 &0.7144  &0.8032  &0.8536 &0.8728  & 0.0992 \\
\bottomrule
\end{tabular}
\end{table*}

\subsection{Results}
\subsubsection{Visual Quality Evaluation}
Tables~\ref{tab:benchmark_part1} and~\ref{tab:benchmark_part2} summarize video quality evaluation results across six evaluation dimensions. 
Overall, embodied world models exhibit stronger performance on structure- and interaction-related metrics, while general-purpose video models mainly excel in perceptual quality. 
Among embodied models, CtrlWorld and TesserAct score highly in subject consistency, background stability, and trajectory accuracy, indicating better alignment with manipulation dynamics. WoW shows strong action-following ability, while RoboMaster and Vidar maintain balanced performance across motion smoothness and content consistency. The open-source video model CogvideoX excels in visual quality and content consistency but lags in physics adherence and motion quality. Closed-source commercial models (Veo 3.1 and Wan 2.6) achieve the highest visual and aesthetic scores, though they show limited improvements in embodied-specific metrics. Qualitative results suggest that visually strong models tend to suffer from semantic drift, while embodied world models produce more coherent and goal-consistent action sequences.

\subsubsection{Embodied Task Evaluation}
\label{sec:embodied_task_eval}
\begin{table}[t]
  \centering
  \caption{Task success rate of downstream policy models trained with generated data from different world models.}
  \label{tab:engine}
  % \resizebox{0.5\textwidth}{!}{
  \begin{tabular}{lcc}
    \toprule
    Model & Task 1 & Task 2  \\
    \midrule
    $\pi_{0.5}$ policy model (zero-shot)     &  2\% & 5\% \\
    $\pi_{0.5}$ policy model (trained with real data)     &  77\% & 66\% \\
    % GigaWorld-0~\cite{team2025gigaworld}      &  &  \\
    Genie Envisioner~\cite{liao2025genie} &7\%  &21\%  \\
    TesserAct~\cite{zhen2025tesseract}        & 1\% & 35\% \\
    RoboMaster~\cite{robomaster_website}       & 7\%  & 68\% \\
    Vidar~\cite{feng2025vidar}            & 13\% & 53\% \\
    WoW~\cite{chi2025wow}              & 45\% & 71\% \\
    Wan 2.2~\cite{wan2025wan}   & 15\% & 41\% \\

    \bottomrule
  \end{tabular}
  % }
\end{table}

\begin{figure}[t]
    \centering   
    \includegraphics[width=0.48\textwidth]{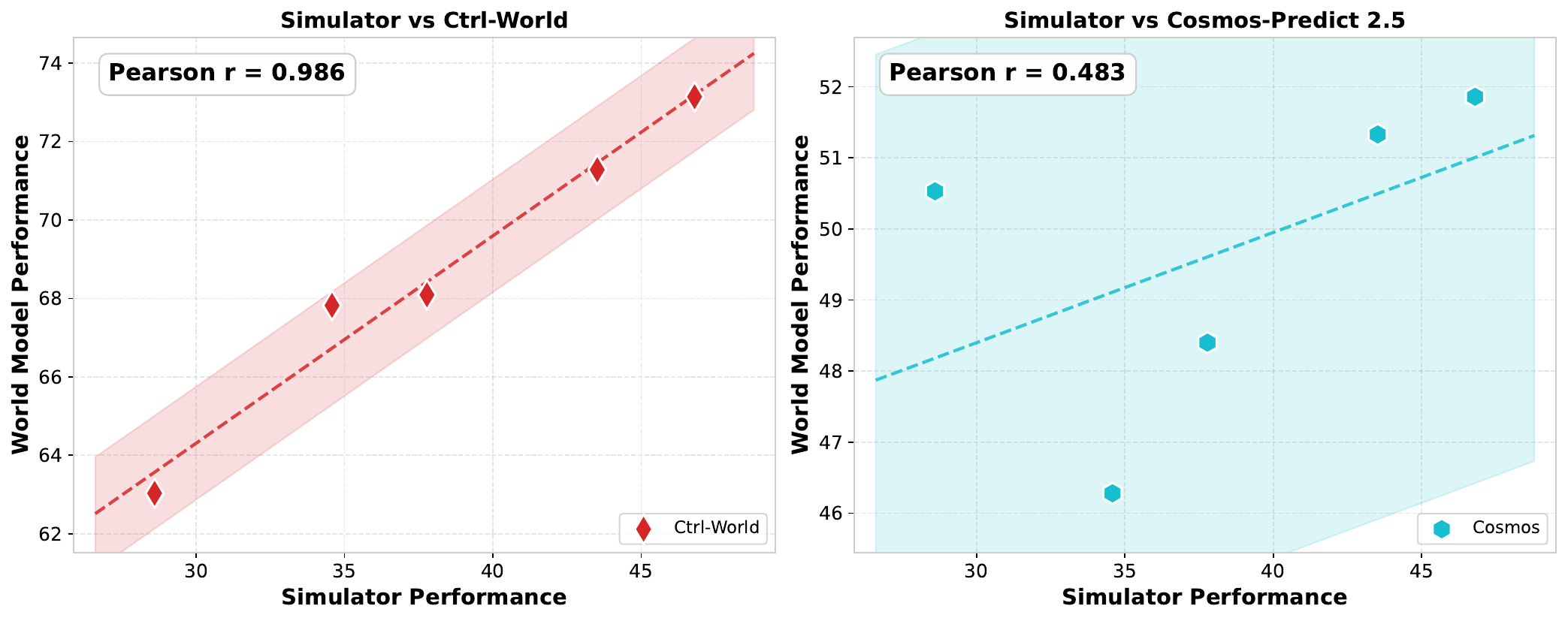} 
    \caption{Correlation of policy evaluation results from world models and the physical simulator.}
    \label{fig:corr_metric}
\end{figure}

\begin{table}[t]
  \centering
  \caption{Task success rate of different world models directly as action planners in the RoboTwin simulator.}
  \label{tab:planner}
  % \resizebox{0.5\textwidth}{!}{
  \begin{tabular}{lcc}
    \toprule
    Model & Task 1 & Task 2  \\
    \midrule
    $\pi_{0.5}$ policy model     & 77\% & 66\% \\
    % GigaWorld-0~\cite{team2025gigaworld}      &  &  \\
    Genie Envisioner~\cite{liao2025genie} & 10\% & 20\% \\
    TesserAct~\cite{zhen2025tesseract}        & 1\% & 35\% \\
    RoboMaster~\cite{robomaster_website}       & 8\% & 20\% \\
    Vidar~\cite{feng2025vidar}            & 2\% & 19\% \\
    WoW~\cite{chi2025wow}              & 20\% & 21\% \\
    Wan 2.2~\cite{wan2025wan}   & 12\% & 20\% \\
    \bottomrule
  \end{tabular}
  % }
\end{table}

\begin{figure*}[t]
    \centering   \includegraphics[width=0.95\textwidth]{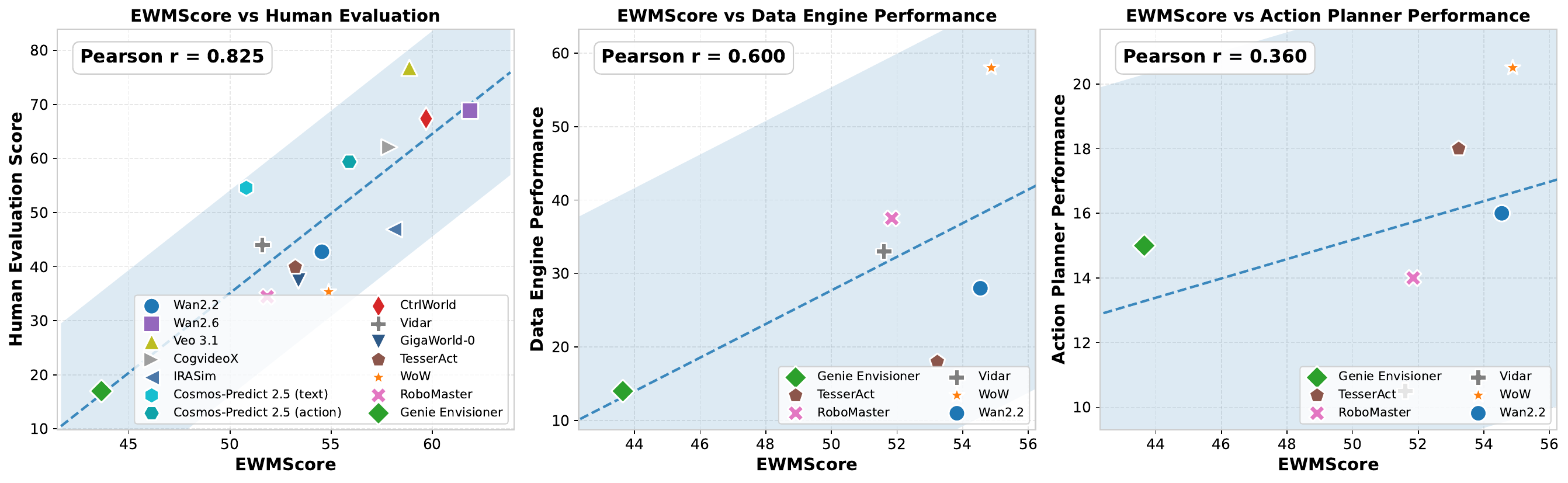} 
    \caption{Correlation between EWMScore with human evaluation and embodied task performance results.}
    \label{fig:corr_all}
\end{figure*}

In this section, we evaluate the capabilities of world models through three embodied tasks.

\textbf{Embodied Data Engine.}
We evaluate six representative world models as data synthesis engines by measuring their impact on downstream policy learning. The evaluation is conducted on two manipulation tasks: \textit{adjust bottle} (Task 1) and \textit{click bell} (Task 2), each executed 100 times, with the success rate averaged. For each task, we train a $\pi_{0.5}$ policy using 25 synthetic trajectories generated by each world model. As shown in Table~\ref{tab:engine}, we observe that synthetic data from most world models provides some performance gains across both tasks but still lags behind real data. Only generated data from RoboMaster and WoW surpass real-data training on Task~2. These results suggest that the quality of generated data remains insufficient for effective policy training, indicating that current embodied world models are not yet reliable data sources for downstream learning.

\textbf{Embodied Policy Evaluator.}
We investigate whether world models can serve as proxy simulation environments for policy evaluation. To this end, we train five policy models $\pi_{0.5}$ with varying performance levels. Each policy is then evaluated by interacting with an action-controllable world model, which generates observation rollouts conditioned on the policy’s actions. 
% Task success is assessed using a VLM-based judge (Qwen 3-VL).
As shown in Figure~\ref{fig:corr_metric}, CtrlWorld exhibits a strong correlation with the evaluation results from the RoboTwin simulator, indicating that it effectively captures meaningful environment transition dynamics. In contrast, Cosmos-Predict 2.5 shows a weaker correlation, suggesting that it struggles to accurately model the environment dynamics. 
Moreover, both models have consistently higher success rates than those measured in the simulator, suggesting partial overfitting to successful trajectories.

\textbf{Embodied Action Planner.}
Similar to the data engine task setting, we evaluate six representative world models as end-to-end action planners by executing their predicted action sequences in the RoboTwin simulator. As shown in Table~\ref{tab:planner}, while several world models achieve non-trivial success rates across tasks, their overall performance remains substantially lower than that of VLA policies such as $\pi_{0.5}$. 
These results indicate that, although current embodied world models capture useful predictive structure, they still struggle to reliably support closed-loop task execution, particularly over long horizons. This indicates significant room for improvement in leveraging world models for autonomous embodied control.

\subsubsection{Human Evaluation}
As shown in Figure~\ref{fig:combined} (b), human evaluations reveal that commercial and large-scale general video models (e.g., Veo 3.1 and Wan 2.6) consistently achieve the highest scores across overall quality, instruction following, and physical adherence, indicating strong perceptual realism and semantic alignment. 
Among embodied world models, action-conditioned approaches such as CtrlWorld demonstrate notably better physical adherence and higher win rates than text-only counterparts, suggesting that explicit action modeling plays a critical role in producing physically plausible interactions. 
In contrast, earlier text-conditioned embodied models (e.g., Genie Envisioner) receive substantially lower scores across all dimensions, reflecting persistent gaps in long-horizon coherence and instruction compliance. 
% Overall, human judgments align well with automatic metrics, which can be seen in Figure~\ref{fig:corr_result}.

\subsection{Inter-metric Analysis}
Figure~\ref{fig:corr_all} presents a cross-dimensional analysis relating EWMScore to both human evaluation and embodied task performance. We observe a strong correlation between EWMScore and human judgments (Pearson $r = 0.825$), indicating a high degree of alignment with subjective perceptual assessments. In contrast, EWMScore exhibits only moderate correlation with data synthesis performance ($r = 0.600$) and a weak correlation with action planning performance ($r = 0.360$). These results suggest that while perceptual realism is a necessary condition for favorable human evaluation, it does not directly translate into proportional gains in downstream embodied tasks. In particular, the limited correlation with action planning indicates that current synthetic data, despite achieving high visual fidelity, remains insufficient to provide strong predictive or decision-relevant signals for complex embodied reasoning.

\section{Conclusion and Future Work}
In this work, we present WorldArena, a unified benchmark for systematically evaluating embodied world models from both perceptual and functional perspectives, integrating multi-dimensional video quality metrics, embodied task evaluations, and human assessments. Through an extensive evaluation of 14 representative models, we reveal consistent gaps between perceptual quality and embodied task performance, highlighting that strong visual generation alone is insufficient for reliable embodied decision-making. We further demonstrate that EWMScore effectively captures overall generative capability and correlates well with human judgments. 
% We hope WorldArena will serve as a foundation for more rigorous, task-aware evaluation and foster the development of embodied world models that are both perceptually strong and functionally reliable.
% We present WorldArena, a unified benchmark for evaluating embodied world models from both perceptual and functional perspectives, integrating video quality metrics, embodied task evaluations, and human assessments. Our evaluation of 11 models reveals a consistent gap between perceptual quality and task performance, emphasizing that strong visual generation alone is not enough for reliable decision-making. We also show that the proposed EWMScore captures generative capability and correlates well with human judgments.
In the future, we will continue to expand WorldArena, incorporating more models to support the advancement of perceptually strong and functionally reliable embodied world models.

\nocite{langley00}

\bibliography{example_paper}
\bibliographystyle{icml2026}

%%%%%%%%%%%%%%%%%%%%%%%%%%%%%%%%%%%%%%%%%%%%%%%%%%%%%%%%%%%%%%%%%%%%%%%%%%%%%%%
%%%%%%%%%%%%%%%%%%%%%%%%%%%%%%%%%%%%%%%%%%%%%%%%%%%%%%%%%%%%%%%%%%%%%%%%%%%%%%%
% APPENDIX
%%%%%%%%%%%%%%%%%%%%%%%%%%%%%%%%%%%%%%%%%%%%%%%%%%%%%%%%%%%%%%%%%%%%%%%%%%%%%%%
%%%%%%%%%%%%%%%%%%%%%%%%%%%%%%%%%%%%%%%%%%%%%%%%%%%%%%%%%%%%%%%%%%%%%%%%%%%%%%%
\newpage
\appendix
\onecolumn

% You can have as much text here as you want. The main body must be at most $8$
% pages long. For the final version, one more page can be added. If you want, you
% can use an appendix like this one.

% The $\mathtt{\backslash onecolumn}$ command above can be kept in place if you
% prefer a one-column appendix, or can be removed if you prefer a two-column
% appendix.  Apart from this possible change, the style (font size, spacing,
% margins, page numbering, etc.) should be kept the same as the main body.
%%%%%%%%%%%%%%%%%%%%%%%%%%%%%%%%%%%%%%%%%%%%%%%%%%%%%%%%%%%%%%%%%%%%%%%%%%%%%%%
%%%%%%%%%%%%%%%%%%%%%%%%%%%%%%%%%%%%%%%%%%%%%%%%%%%%%%%%%%%%%%%%%%%%%%%%%%%%%%%
\section{Additional Details on Metrics} \label{appendix:metric}

\subsection{Image Quality}
\label{appendix:image quality}
The per-frame sharpness of a generated video constitutes the foundation of its visual presentation. Unlike traditional reference-based metrics such as PSNR, which rely on ground-truth images, we adopt the \textbf{MUSIQ} (Multi-scale Image Quality Transformer) model~\cite{ke2021musiq} to evaluate technical distortions in a no-reference setting, including overexposure, sensor noise, and compression artifacts. MUSIQ leverages a multi-scale Transformer architecture to capture the relationships between local details and global composition.

For a video sequence $V = \{I_1, I_2, \dots, I_T\}$,where $V$represents a specific video and $I_i$ represents the $i$th frame of the video $V$, the image quality score $S_{\text{img}}$ is defined as:
\begin{equation}
S_{\text{img}} = \frac{1}{T} \sum_{t=1}^{T} \Phi_{\text{musiq}}(I_t)
\end{equation}
where $\Phi_{\text{musiq}}(\cdot)$ denotes the pretrained quality prediction function. A higher value of $S_{\text{img}}$ indicates greater visual purity and clarity at the level of digital image~\cite{huang2024vbench}.

\subsection{Aesthetic Quality}
\label{appendix:aesthetic quality}
Beyond technical fidelity, generated videos are also required to conform to human aesthetic principles, such as harmonious lighting and visually pleasing color composition. We employ the \textbf{LAION Aesthetic Predictor}~\cite{laion2022aesthetic} to perform aesthetic feature mapping for each frame. Similarly, for a video sequence $V = \{I_1, I_2, \dots, I_T\}$, the aesthetic quality score $S_{\text{aes}}$ is defined as:
\begin{equation}
S_{\text{aes}} = \frac{1}{T} \sum_{t=1}^{T} \Psi_{\text{aes}}(I_t)
\end{equation}
where $\Psi_{\text{aes}}(\cdot)$ maps each image into a high-dimensional feature space and predicts an aesthetic score. This formulation ensures that the evaluation extends beyond pixel-level sharpness to encompass perceptual coherence and artistic consistency~\cite{huang2024vbench}.

\subsection{JEPA Similarity}
\label{appendix:JEPA similarity}
To evaluate video quality from a global feature-distribution perspective and detect high-level spatiotemporal collapse, we introduce the \textbf{JEPA Similarity}. Unlike traditional FVD metric, which relies on Gaussian assumptions, JEPA measures the maximum mean discrepancy (MMD) between feature distributions to provide evaluation results that better align with human perception:
\begin{equation}
S_{\text{JEPA}} = \exp\left(-\alpha \cdot
\widehat{\text{MMD}}^2_{\text{poly}}
(\mathcal{F}_{\text{gen}}, \mathcal{F}_{\text{ref}})\right)
\end{equation}
where $\alpha = 40$ is a scaling factor that enhances numerical distinguishability, $\mathcal{F}_{\text{gen}}$ and $\mathcal{F}_{\text{ref}}$denote the feature space distributions of the generated video set and the reference expert demonstration(GT) set, respectively, extracted by a pretrained \textbf{V-JEPA} encoder~\cite{bardes2023v} which
is pre-trained via masked prediction tasks and enables the model to capture high‑level spatio‑temporal causality and physical logic in videos, offering greater robustness to temporal warping and content variations.$\widehat{\text{MMD}}^2_{\text{poly}}$represents the squared estimator of the Maximum Mean Discrepancy using a second‑order polynomial kernel, defined as $k(\mathbf{x}, \mathbf{y}) = (\gamma \langle \mathbf{x}, \mathbf{y} \rangle + c_0)^2$,with $\gamma=1,c_0=0$, It measures the distance between the two feature sets in the reproducing kernel Hilbert space (RKHS), computed as follows:

\begin{equation}
\widehat{\text{MMD}}^2_{\text{poly}}(\mathcal{F}_{\text{gen}}, \mathcal{F}_{\text{ref}}) = \frac{1}{m(m-1)} \sum_{i \neq j}^{m} k(\mathbf{f}_i^{\text{gen}}, \mathbf{f}_j^{\text{gen}}) + \frac{1}{n(n-1)} \sum_{i \neq j}^{n} k(\mathbf{f}_i^{\text{ref}}, \mathbf{f}_j^{\text{ref}}) - \frac{2}{mn} \sum_{i=1}^{m} \sum_{j=1}^{n} k(\mathbf{f}_i^{\text{gen}}, \mathbf{f}_j^{\text{ref}}),
\end{equation}
where $m$ and $n$ denote the number of samples in the generated videos and the reference videos,$\mathbf{f}_i^{\text{gen}}$ and $\mathbf{f}_i^{\text{ref}}$ are the corresponding V‑JEPA feature vectors. Higher values indicate closer alignment to reference demonstrations~\cite{luo2024jedi}. 

This metric is not only sensitive to breakdowns in high‑level spatio‑temporal logic but also avoids the Gaussian distribution assumption, offering significantly better sample efficiency than conventional metrics. Moreover, it achieves a enormous improvement in correlation with human subjective assessments, especially when evaluating complex embodied operation logic, thereby more reliably reflecting the physical plausibility and spatio‑temporal consistency of the generated videos.

\subsection{Dynamic Degree}
\label{subsec:dynamic}
We employ the \textbf{RAFT}(Recurrent All-Pairs Field Transforms)~\cite{teed2020raft} optical flow model to extract motion vector fields between adjacent frames. To accurately capture the most representative and salient motions in a video,such as robotic arm grasping,we focus on pixels whose optical flow magnitudes fall within the top $5\%$~\cite{huang2024vbench}.

Let $\mathbf{u}_{t,t+1}$ denote the two-dimensional optical flow field between consecutive frames. We define the average magnitude of the active pixel set as $\bar{v}_{\text{top5}}$. To obtain a smooth numerical mapping while introducing resolution adaptivity, we define the dynamic degree score as:
\begin{equation}
S_{\text{dyn}} = \frac{1}{1 + \exp\left( -\alpha \cdot \left(\frac{\bar{v}_{\text{top5}}}{\tau} - 1\right) \right)}
\end{equation}
where $\tau = \frac{6}{256} \times \min(H, W)$ is a resolution-adaptive threshold constant, $\alpha$ controls the steepness of the mapping curve, and $S_{\text{dyn}} \in (0, 1)$. Values closer to $1$ indicate more pronounced dynamic responses in the video.

\subsection{Flow Score}
\label{appendix:flow score}
To quantify overall physical motion intensity and dynamic activity, we compute an optical-flow-based motion score. Given a generated video $V = \{I_1, \dots, I_T\}$ with frame width $W$ and height $H$, we use \textbf{RAFT}~\cite{teed2020raft} to estimate dense optical flow fields $\mathbf{u}_t \in \mathbb{R}^{H \times W \times 2}$ between consecutive frames $I_t$ and $I_{t+1}$. By averaging the magnitude of optical flow across all pixels,the average motion intensity is defined as:
\begin{equation}
S_{\text{flow\_raw}} =
\frac{1}{T-1}
\sum_{t=1}^{T-1}
\left(
\frac{1}{H \cdot W}
\sum_{i,j}
\lVert \mathbf{u}_t(i,j) \rVert_2
\right)
\end{equation}
where $(i,j)$ indexes pixel locations,$\|\cdot\|_2$ denotes the Euclidean norm ($L_2$ norm), which quantifies the magnitude of pixel displacement per unit time~\cite{liu2023evalcrafter}. In the context of embodied intelligence tasks, this metric serves a dual evaluative purpose: on the one hand, it effectively identifies whether a video degenerates into "static frames" or exhibits only "minimal drift" due to insufficient generative capability of the model; on the other hand, it captures whether unnatural, non‑physical distortions are present in the overall scene. 

Higher values of $S_{\text{flow\_raw}}$ typically indicate more pronounced dynamic interaction and physically meaningful motion. Compared to \textbf{Dynamic Degree}, this metric focuses on assessing overall motion intensity and detecting implausible global dynamics. To ensure consistent interpretation and comparability with other metrics, we will normalize $S_{\text{flow\_raw}}$ to the range $[0, 1]$ in Section~\ref{subsec:normalization}, denoting the normalized value as $S_{\text{flow}}$, while preserving the property that higher values correspond to better performance.

\subsection{Motion Smoothness}
\label{appendix:motion smoothness}
To evaluate whether motion is temporally coherent and consistent with physical inertia, we adopt a reconstruction-based strategy using a video frame interpolation model (\textbf{VFI-Mamba})\cite{zhang2024vfimambavideoframeinterpolation}. Given all frames of a video, we take the odd-indexed frames $\{I_1, I_3, \dots\}$ as inputs and predict the corresponding intermediate frames $I_{\text{mid}}$, which are then compared with the ground-truth(GT) frames. If the motion is physically plausible and smooth, the intermediate frame $I_{\text{mid}}$ should be accurately reconstructed from its surrounding frames $(I_{\text{prev}}, I_{\text{next}})$ via nonlinear interpolation.

The key innovation lies in incorporating motion magnitude as a weighting factor to avoid overestimating static backgrounds. The final motion smoothness score is defined as:
\begin{equation}
S_{\text{smooth\_raw}} =
\frac{1}{N}
\sum
\text{SSIM}(\hat{I}_{\text{pred}}, I_{\text{mid}})
\cdot
\ln\left(1 + \text{diff}(I_{\text{prev}}, I_{\text{next}})\right)
\end{equation}
where $N$ denotes the number of predicted intermediate frames (typically equal to or one less than the number of even-indexed frames), $\hat{I}_{\text{pred}}$ is the interpolated frame predicted by the model,$I_{\text{mid}}$ is the real frame between $I_{\text{prev}}$and$I_{\text{next}}$, and $\text{diff}(\cdot)$ represents the mean raw pixel-wise difference between two frames. The logarithmic weighting $\ln(1+x)$ compensates for the increased difficulty of interpolation under large motion, thereby assigning higher rewards to sequences that maintain high reconstruction fidelity even during rapid motion. For consistency with other evaluation metrics, $S_{\text{smooth\_raw}}$ will be normalized to the range $[0, 1]$ in Section~\ref{subsec:normalization}, yielding the final smoothness score $S_{\text{smooth}}$, where higher values indicate superior temporal coherence and motion consistency.

\subsection{Subject Consistency}
\label{appendix:subject consistency}
For a video sequence $V = \{I_1, I_2, \dots, I_T\}$, we extract frame-level features using \textbf{DINO}~\cite{caron2021emerging}, denoted as $f_i = \text{DINO}(I_i)$, which emphasize the spatial topological structure of objects. We compute the cosine similarity between the feature $f_i$ of the current frame and both the first-frame feature $f_1$ and the previous-frame feature $f_{i-1}$, and average the similarities across all frames~\cite{huang2024vbench}:
\begin{equation}
    S_{\text{subj\_raw}} = \sum_{t=2}^{T} \left(\frac{  \cos(f_i, f_1)  +  \cos(f_i, f_{i-1})}{2} \right)
\end{equation}

However, a common “shortcut” phenomenon in video generation evaluation is that models may produce nearly static videos to obtain artificially high consistency scores. To faithfully reflect dynamic generation capability in embodied scenarios, we introduce the dynamic degree $S_{\text{dyn}}$ defined in Section~\ref{subsec:dynamic} as a weighting factor for subject consistency. Specifically, when the video’s dynamic degree falls below a predefined threshold $\gamma$, the raw score is penalized as:
\begin{equation}
    S_{\text{subj}} = S_{\text{subj\_raw}} \cdot \min(1, \frac{S_{\text{dyn}}}{\gamma})
\end{equation}

This mechanism ensures that static or near-static videos cannot achieve high scores even when frame-level similarity is extremely high, leading to more reasonable evaluation in embodied tasks.

\subsection{Background Consistency}
\label{appendix:background consisitency}
Analogous to subject consistency, for a video sequence $V = \{I_1, I_2, \dots, I_T\}$, we extract frame-level features using \textbf{CLIP}~\cite{radford2021learningtransferablevisualmodels}, denoted as $h_i = \text{CLIP}(I_i)$, which place greater emphasis on global scene semantics and prevent uncontrolled background variation during generation. We compute the cosine similarity between $h_i$ and both $h_1$ and $h_{i-1}$, and average the results across all frames~\cite{huang2024vbench}:
\begin{equation}
S_{\text{bg\_raw}} = \sum_{t=2}^{T} \left( \frac{\cos(h_i, h_1) + \cos(h_i, h_{i-1})}{2} \right)
\end{equation}
Similarly, the final background consistency score is adjusted using the dynamic degree:
\begin{equation}
S_{\text{bg}} = S_{\text{bg\_raw}} \cdot \min\left(1, \frac{S_{\text{dyn}}}{\gamma}\right)
\end{equation}

\subsection{Photometric Consistency}
\label{appendix:photometric consistency}
Photometric consistency measures the physical stability of textures at the pixel level. For a video $V = \{I_1, I_2, \dots, I_T\}$, we use the forward optical flow field $\mathbf{u}_t$ between frames $I_t$ and $I_{t+1}$, as well as the backward flow field $\mathbf{u}'_{t+1}$, to warp pixels from frame $t$ to frame $t+1$ and then back to frame $t$. The average end-point error (AEPE) is defined as~\cite{duan2025worldscore}:
\begin{equation}
    E_{\text{photo}} =\frac{1}{T}\sum_{t=1}^{T} \| \text{Warp}_{back}(\text{Warp}_{fwd}(I_t, \mathbf{u_{t}}), \mathbf{u'_{t+1}}) - I_t \|_2
\end{equation}
Since this metric quantifies pixel-level reconstruction error, lower values correspond to superior visual quality. To obtain a positively correlated measure that appropriately rewards sequences with meaningful motion while penalizing trivial solutions in static videos, we compute the pre-normalized photometric consistency score as:
\begin{equation}
S_{\text{photo\_raw}} = \frac{1}{E_{\text{photo}}} \cdot \min\left(1, \frac{S_{\text{dyn}}}{\gamma}\right)
\end{equation}
where $S_{\text{dyn}}$ denotes the dynamic degree (formally defined in Section~\ref{subsec:dynamic}), quantifying the overall motion intensity within a video sequence, and $\gamma$ serves as a dynamic threshold that modulates the penalty for insufficient motion. The inclusion of $S_{\text{dyn}}$ addresses a critical limitation of conventional photometric metrics: static or near-static sequences often achieve artificially high scores due to minimal frame-to-frame variations, even though they fail to demonstrate meaningful dynamic modeling. 

By scaling the raw reciprocal score with the normalized dynamic degree, our formulation ensures that only videos with sufficient motion ($S_{\text{dyn}} \geq \gamma$) retain their full photometric consistency score, while static sequences are proportionally penalized. This encourages the model to maintain high reconstruction fidelity under actual motion rather than exploiting static scenarios. Subsequently, $S_{\text{photo\_raw}}$ is normalized to the interval $[0, 1]$ in Section~\ref{subsec:normalization}, which produces the final photometric consistency metric $S_{\text{photo}}$, where higher values denote enhanced visual fidelity and temporal coherence.

\subsection{Interaction Quality}
\label{appendix:interaction quality}
This metric evaluates the physical plausibility of interactions between the robotic arm and environmental objects, including contact behavior, force transmission, friction, inertia, and boundary integrity. We employ the pretrained multimodal model \textbf{Qwen3-VL-8B}~\cite{Qwen3-VL} as a VLM-based judge. Given $N_{\text{sample}}$ sampled frames and the task instruction, the model assigns a 1–5 Likert score, which is normalized to $[0,1]$ to yield the final interaction quality score,the prompt used to evaluate interaction quality can be found in \ref{tab:VLM_Evaluation_System_Prompt}.

\begin{tcolorbox}[
    colback=gray!5!white,
    colframe=gray!75!black,
    title=VLM Evaluation Prompt,
    fonttitle=\bfseries,
    left=5pt, right=5pt, top=5pt, bottom=5pt,
    sharp corners,
    boxrule=0.5pt,
    breakable % 如果内容过长，允许跨页
]
\label{tab:VLM_Evaluation_System_Prompt}
\small
\noindent
You are an expert evaluator for robot interaction videos. You are evaluating videos generated for **embodied AI manipulation scenarios**, specifically focusing on robotic arms interacting with objects in tabletop environments.

\vspace{0.5em}
\noindent
**EVALUATION CONTEXT:** \\
- Target scenario: Robotic manipulation (e.g., pick-place, push, grasp) \\
- Expected agent: **Robotic arm/end-effector**, NOT human hands \\
- Expected environment: Tabletop with objects, typical for robot manipulation tasks \\
- Expected physics: Realistic robot-object interactions following physical laws

\vspace{0.5em}
\noindent
**CRITICAL EVALUATION PRINCIPLES:** \\
1. Base ALL judgments ONLY on what is visually observable in the sampled frames \\
2. DO NOT infer information not shown (no assumptions about unseen parts) \\
3. Evaluate temporal coherence across the sampled frames \\
4. For instruction following: Compare STRICTLY against the provided text instruction

\vspace{0.5em}
\noindent
**EVALUATION DIMENSIONS \& SCORING RUBRICS:**

\vspace{0.5em}
\noindent
1. Interaction\_Quality (Quality of robot-object interactions) \\
\indent - Score 1: Objects pass through robot or other objects; no proper contact \\
\indent - Score 2: Contact exists but interaction is unrealistic (e.g., sliding without friction, incorrect force response) \\
\indent - Score 3: Mostly plausible interactions with minor issues (e.g., slight penetration, imperfect grasping) \\
\indent - Score 4: Realistic contact physics (proper friction, force transfer, object deformation) \\
\indent - Score 5: Perfect interaction physics; indistinguishable from real robot manipulation

\vspace{0.5em}
\noindent
2. PERSPECTIVITY (3D consistency and camera geometry) \\
\indent - Score 1: Scene has no coherent 3D structure; objects float inconsistently \\
\indent - Score 2: 3D structure is unstable (e.g., scale changes, incorrect occlusion) \\
\indent - Score 3: Reasonable 3D consistency with minor issues (e.g., slight perspective drift) \\
\indent - Score 4: Stable camera perspective with consistent depth relationships \\
\indent - Score 5: Perfect camera geometry and 3D consistency

\vspace{0.5em}
\noindent
3. INSTRUCTION FOLLOWING (Adherence to given instruction:**VIDEO INSTRUCTION**) \\
\indent - **HALLUCINATION CHECK**: If the video shows human hands instead of robotic arms, score $\le$ 2 immediately \\
\indent - Score 1: Completely different from instruction (wrong action, wrong objects, wrong scene) \\
\indent - Score 2: Partially related but major errors (e.g., wrong target object, incorrect manipulation type) \\
\indent - Score 3: Follows general intent but with execution errors (e.g., correct action sequence but imprecise) \\
\indent - Score 4: Mostly correct with minor deviations (e.g., slight position error, extra unnecessary motion) \\
\indent - Score 5: Perfect execution of all specified elements (action, object, scene, outcome)

\vspace{0.5em}
\noindent
**SPECIFIC ROBOT-RELATED CHECKS:** \\
- Robotic arm should have mechanical appearance, NOT human limbs \\
- End-effector (gripper) should maintain consistent form throughout interaction \\
- Robot motion should show appropriate joint movement and kinematics \\
- Object manipulation should respect object mass and inertia \\
- Contact should be maintained appropriately during grasping/lifting

\vspace{0.5em}
\noindent
**OUTPUT FORMAT REQUIREMENTS:** \\
You MUST output a SINGLE, VALID JSON object with EXACTLY three keys: \\
- 'Interaction\_Quality' \\
- 'Perspectivity' \\
- 'Instruction\_Following'

\vspace{0.5em}
\noindent
Each value must be an object with exactly two keys: \\
- `"score"`: integer 1-5 \\
- `"reason"`: concise explanation citing SPECIFIC visual evidence from frames

\vspace{0.5em}
\noindent
**EXAMPLE OUTPUT:**
\begin{verbatim}
{
  "Interaction_Quality": {
    "score": 2,
    "reason": "Object slides without friction during pushing; 
               gripper penetrates object slightly"
  },
  "Perspectivity": {
    "score": 4,
    "reason": "Stable camera perspective with consistent depth ordering"
  },
  "Instruction_Following": {
    "score": 1,
    "reason": "Video shows human hand instead of robotic arm (hallucination)"
  }
}
\end{verbatim}

\vspace{0.5em}
\noindent
**CRITICAL INSTRUCTIONS:** \\
1. Output ONLY the JSON object, no other text \\
2. Base scoring on observed visual evidence only \\
3. For instruction following: Strictly compare with the provided instruction \\
4. Consider temporal coherence across all sampled frames \\
5. Penalize hallucinations (e.g., human hands instead of robot) heavily

\vspace{0.5em}
\noindent
Now evaluate the provided video frames based on the above criteria.
\end{tcolorbox}
\subsection{Trajectory Accuracy}
\label{appendix:trajectory quality}
In embodied intelligence tasks, the accuracy of the robotic arm’s grasping trajectory is a core indicator of whether the model generates \emph{effective actions}. Trajectories encode not only low-level physical consistency but also high-level task logic and interaction constraints. To quantify this property, we first apply \textbf{SAM3} (Segment Anything Model 3)~\cite{carion2025sam} to extract bounding boxes of the robotic arm in each frame. After non-maximum suppression(nms) and confidence filtering, we construct the raw trajectory sequences using the centers of candidate boxes.

Let the ground-truth trajectory be $GT = (r_1, r_2, \dots, r_{|R|})$ and the generated trajectory be $P = (p_1, p_2, \dots, p_{|P|})$, where $|R|$ and $|P|$ denote the sequence lengths, respectively. To address missing detections caused by occlusion or tracking interruption, we apply linear interpolation to ensure temporal continuity. For a missing point $p_i$ with $i \notin M$, its position is computed as:
\begin{equation}
p_i = (1 - \alpha) p_{\text{prev}} + \alpha p_{\text{next}},
\quad
\alpha = \frac{i - \text{prev}}{\text{next} - \text{prev}}
\end{equation}
where $\text{prev}$ and $\text{next}$ denote the nearest valid observation indices before and after $i$.

We then compute the normalized dynamic time warping distance (NDTW)~\cite{book} to evaluate global alignment between the generated trajectory and the ground-truth trajectory:
\begin{equation}
\text{NDTW}(GT, P) =
\min_{\pi}
\frac{1}{|R|}
\sqrt{
\sum_{(i,j) \in \pi}
\lVert r_i - p_j \rVert^2
}
\end{equation}
where $\pi$ denotes the optimal alignment path. This metric captures both temporal causality and task-stage ordering, enabling discrimination between correct and incorrect execution sequences such as "approach-grasp-move." Since lower NDTW values indicate better alignment, we first derive a pre-normalized trajectory alignment score~\cite{yue2025ewmbench}:
\begin{equation}
S_{\text{traj\_raw}} = \frac{1}{\text{NDTW}(GT, P)}
\end{equation}
where higher values correspond to more accurate spatial-temporal alignment with the real trajectory and more accurate actions. To ensure consistency with our evaluation framework and facilitate direct comparison with other metrics, we normalize $S_{\text{traj\_raw}}$ to the range $[0, 1]$ in Section~\ref{subsec:normalization}, yielding the final trajectory alignment score $S_{\text{traj}}$. This normalized metric preserves the property that higher values indicate superior trajectory fidelity and task-stage adherence.

\subsection{Depth Accuracy}
\label{appendix:depth quality}
To evaluate whether the generated video preserves real-world spatial geometry, we compute depth discrepancies between the generated video and the ground-truth reference using the monocular depth estimation model \textbf{Depth-Anything}~\cite{depth_anything_v2}. Since monocular depth prediction suffers from scale ambiguity, we adopt a \textbf{median-based scaling} strategy~\cite{worldlens}.

The procedure is as follows:
\begin{enumerate}
\item \textbf{Uniform Sampling}: We uniformly sample $T_{\text{target}} = 40$ frames from both the generated video and the ground-truth video to ensure temporal alignment.
\item \textbf{Scale Alignment}: Depth maps $D_{\text{gen}}$ and $D_{\text{gt}}$ are estimated for the generated and ground-truth frames, respectively. Their medians are computed as $m_{\text{gen}} = \text{median}(D_{\text{gen}})$ and $m_{\text{gt}} = \text{median}(D_{\text{gt}})$. The scaling factor $\alpha = \frac{m_{\text{gt}}}{m_{\text{gen}}}$ is applied to obtain the aligned depth $\hat{D}_{\text{gen}} = D_{\text{gen}} \cdot \alpha$.
\item \textbf{AbsRel Error}: Within the valid pixel mask $\mathcal{M}$ (which typically filters out noise and distant regions with ground‑truth depth $D_{\text{gt}}<1e-3$), the absolute relative error is computed as follows:
\end{enumerate}
\begin{equation}
E_{\text{Depth}} =
\frac{1}{|\mathcal{M}|}
\sum_{p \in \mathcal{M}}
\frac{|\hat{D}_{\text{gen}}(p) - D_{\text{gt}}(p)|}{D_{\text{gt}}(p) + \epsilon}
\end{equation}
where $\epsilon$ is a small constant to prevent division by zero. Lower values indicate stronger depth accuracy with the real-world scene. To align this metric with our evaluation framework where higher scores correspond to better performance, we will normalize $E_{\text{Depth}}$ to the range $[0, 1]$ and invert its direction in Section~\ref{subsec:normalization},such that higher values correspond to higher accuracy, resulting in the final normalized depth accuracy score $S_{\text{Depth}}$.

\subsection{Perspectivity}
\label{appendix:perspectivity}
This metric evaluates three-dimensional geometric plausibility. The VLM examines perspective cues such as scale variation with depth, lighting consistency, and occlusion relationships during camera motion. We use \textbf{Qwen3-VL-8B} as a judge and normalize the Likert-scale output to $[0,1]$,which is normalized to $[0,1]$ to yield the final perspectivity score,the prompt used to evaluate perspectivity can be found in \ref{tab:VLM_Evaluation_System_Prompt}.

\subsection{Instruction Following}
\label{appendix:instruction following}
This metric evaluates the semantic consistency between each generated video $V_i$ and its corresponding instruction $Inst_i$, focusing on action type, target object, and final task state. We again use \textbf{Qwen3-VL-8B} as a VLM-based judge with a normalized 1–5 Likert scale, which is normalized to $[0,1]$ to yield the final instruction following score,the prompt used to evaluate instruction following can be found in \ref{tab:VLM_Evaluation_System_Prompt}.

\subsection{Semantic Alignment}
\label{appendix:semantic alignment}
To assess whether the generated video truly understands and executes the given textual instruction, we evaluate semantic alignment as follows:
\begin{equation}
S_{\text{clip}} =
w \cdot
\max\left( \cos(f_{\text{gen}}, f_{\text{gt}}), 0 \right)
\end{equation}
where $f_{\text{gen}} \in \mathbb{R}^d$ denotes the semantic feature vector of the generated video. Specifically, we first employ a vision–language model (VLM), \textbf{Qwen2.5-VL}~\cite{Qwen2.5-VL}, to produce a dense structured description $L_{\text{gen}}$ of the generated video under task-oriented prompting, covering both task summary and action sequence. This text is then encoded by the CLIP~\cite{radford2021learningtransferablevisualmodels} text encoder $\Phi_{\text{txt}}$, yielding $f_{\text{gen}} = \Phi_{\text{txt}}(L_{\text{gen}})$.

Similarly, $f_{\text{gt}} \in \mathbb{R}^d$ denotes the semantic feature vector of the ground-truth(GT) video, obtained via the same pipeline as $f_{\text{gt}} = \Phi_{\text{txt}}(L_{\text{gt}})$, where $L_{\text{gt}}$ is the structured description of the reference video. The scaling factor $w$ ensures score normalization. A higher value of $S_{\text{clip}}$ indicates stronger semantic alignment between the generated video and the reference execution.

\subsection{Action Following}
\label{appendix:action following}
This metric evaluates the model’s ability to produce distinct and correct outcomes for different action instructions. In open-loop prediction tasks, a robust model should execute multiple instructions faithfully rather than collapsing into repetitive patterns. Given a single action instruction, we manually annotate or automatically generate multiple distinct action instructions and prompt the model to generate $N$ corresponding videos.

For each generated video $V_k$, we extract a global CLIP feature vector $f_k$. The action-following diversity score is computed as the average pairwise feature dissimilarity(1-cosine similarity between two vectors$f_i$and$f_j$~\cite{yue2025ewmbench}:
\begin{equation}
S_{\text{div}} =
\frac{1}{|\text{Pairs}(i,j)|}
\sum_{i < j}
\left(
1 -
\frac{f_i \cdot f_j}{\lVert f_i \rVert \lVert f_j \rVert}
\right)
\end{equation}
A higher value of this metric indicates stronger capability of the model in correctly executing action instructions,which is already normalized.

\subsection{Score Normalization}
\label{subsec:normalization}
Several metrics in our evaluation framework require normalization and direction alignment to ensure consistent interpretation and fair comparison across different models. Specifically, the \textbf{Flow Score}, \textbf{Trajectory Accuracy}, \textbf{Photometric Consistency}, and \textbf{Motion Smoothness} metrics are initially measured on different scales, while some metrics such as \textbf{JEPA Similarity} and \textbf{Depth Accuracy} represent error measures where lower values indicate better performance. To address these inconsistencies, we apply a two-step normalization procedure.

For the \textbf{Flow Score}, \textbf{Trajectory Accuracy}, \textbf{Photometric Consistency}, and \textbf{Motion Smoothness}  metrics, we employ \textbf{empirical min-max} normalization based on the distribution of scores across all evaluated models. We compute the $99^{\text{th}}$ and $1^{\text{st}}$ percentiles of each metric across all videos generated by the 8 models, which serve as the empirical maximum and minimum bounds, respectively.And the specific numerical values for these empirical bounds are provided in Table~\ref{tab:empirical_bounds}. The final normalized score is calculated as:

\begin{equation}
S_{\text{final}} = \max\left(0, \min\left(1, \frac{S_{\text{raw}} - S_{\text{empirical}}^{\text{min}}}{S_{\text{empirical}}^{\text{max}} - S_{\text{empirical}}^{\text{min}}}\right)\right)
\end{equation}

where $S_{\text{raw}}$ denotes the raw metric value, $S_{\text{empirical}}^{\text{max}}$ and $S_{\text{empirical}}^{\text{min}}$ represent the empirical bounds. This transformation ensures that all scores reside within the interval $[0, 1]$, with higher values indicating better performance.

For \textbf{Depth Accuracy}, which originally measures reconstruction error (lower values are better), we apply the same normalization but invert the direction:

\begin{equation}
S_{\text{final}} = 1 - \max\left(0, \min\left(1, \frac{S_{\text{raw}} - S_{\text{empirical}}^{\text{min}}}{S_{\text{empirical}}^{\text{max}} - S_{\text{empirical}}^{\text{min}}}\right)\right)
\end{equation}

\begin{table}[htbp]
\centering
\caption{Empirical bounds for metric normalization. The values represent the $99^{\text{th}}$ percentile (maximum) and $1^{\text{st}}$ percentile (minimum) of each metric across all evaluated videos.}
\label{tab:empirical_bounds}
\begin{tabular}{lcc}
\toprule
\textbf{Metric} & \textbf{Empirical Maximum} ($S_{\text{empirical}}^{\text{max}}$) & \textbf{Empirical Minimum} ($S_{\text{empirical}}^{\text{min}}$) \\
\midrule
Photometric Consistency\hfill \textbf{(Higher is Better)} & 6.7899 & 0.1257\\
Motion Smoothness    \hfill \textbf{(Higher is Better)} & 2.6413 & 0.0000 \\
Trajectory Accuracy \hfill \textbf{(Higher is Better)} & 40.8540 & 0.0000 \\
Flow Score \hfill \textbf{(Higher is Better)} & 8.9414 & 0.0531 \\
Depth Accuracy \hfill \textbf{(Lower is Better)} & 4.3711 & 0.2228 \\
\bottomrule
\end{tabular}
\end{table}

This comprehensive normalization strategy ensures that all metrics are scaled to the unit interval $[0, 1]$, aligned in direction (higher values always denote better performance), and comparable across different evaluation dimensions.
% \subsection{ Empirical Bounds}
% In this section, we discuss how we calculate the empirical
% bounds for each evaluation dimension, which will be used
% for linear normalization in
% \subsection{ Score Normalization}
% \label{subsec:normalization}
% A higher value of $S_{\text{div}}$ indicates stronger capability in correctly following diverse action instructions.
\newpage
\section{The Prompt of VLM-based Policy Success Judgement in Policy Evaluator Task}
\label{appendix:policy_judge_prompt}

In the embodied policy evaluator task (Section~\ref{sec:embodied_task_eval}), we assess whether world models can serve as proxy simulation environments for policy evaluation. To determine task success, we employ a VLM-based judge that compares the policy-generated video rollouts against ground-truth reference trajectories. The judge evaluates three critical aspects: (1) correct arm selection when specified in the instruction, (2) task completion by comparing final states between generated and ground-truth videos, and (3) overall action intent consistency. This evaluation approach accounts for visual artifacts inherent to world model rendering while focusing on functional correctness, enabling scalable and automated assessment of policy execution quality. The complete system prompt used for this VLM-based evaluation is provided below.

\begin{tcolorbox}[
  colback=gray!5!white,
  colframe=gray!75!black,
  title=Policy Execution Evaluation System Prompt,
  fonttitle=\bfseries,
  breakable
]
\label{tab:Policy_Judge_System_Prompt}
\small
\noindent
You are a robot task execution judge. Please determine if the policy model correctly executed the instruction.

\vspace{0.5em}
\noindent
**Task Instruction**: \{instruction\}

\vspace{0.5em}
\noindent
**INPUT DESCRIPTION:** \\
- First 5 images: GT (Ground Truth) video frames (uniformly sampled: first frame, 3 middle frames, last frame), showing the correct task execution \\
- Last 5 images: Policy model generated video frames, showing the policy model's execution

\vspace{0.5em}
\noindent
**EVALUATION CRITERIA** (by priority):

\vspace{0.5em}
\noindent
1. Arm Selection - If the instruction explicitly requires left/right arm, the correct arm must be used, otherwise fail \\
2. Task Completion - Compare GT's final state with Policy's final state: \\
\indent - GT's final frame shows the completed task state \\
\indent - Policy's final frame should show a similar completion state \\
\indent - If Policy's final frame differs significantly from GT's final frame, judge as failure \\
3. Action Intent - Is Policy's entire motion process consistent with the instruction's semantic meaning?

\vspace{0.5em}
\noindent
**TOLERABLE DIFFERENCES:** \\
- Visual hallucinations from world model rendering (object deformation, color shifts) \\
- Minor differences in action trajectory \\
- Video length differences

\vspace{0.5em}
\noindent
**JUDGE AS SUCCESS (1):** \\
- Correct arm used \\
- Final state similar to GT (task basically completed) \\
- Correct action intent

\vspace{0.5em}
\noindent
**JUDGE AS FAILURE (0):** \\
- Wrong arm used \\
- Final state significantly different from GT (task not completed or completed incorrectly) \\
- Completely wrong action direction \\
- Grabbed/operated wrong object

\vspace{0.5em}
\noindent
Please carefully compare the **final frames** of GT and Policy to judge if the task is basically completed.

\vspace{0.5em}
\noindent
**OUTPUT FORMAT REQUIREMENTS:** \\
Please respond in this format: \\
thinking: [Analysis: 1. Is arm correct? 2. Compare final frame task completion states 3. Is action intent consistent?] \\
answer: [0 or 1]
\end{tcolorbox}

\section{Case Comparison of Each Metric in EWMScore}
\label{appendix:metric visualization}

\begin{figure*}[h!]
    \centering   \includegraphics[width=0.6\textwidth]{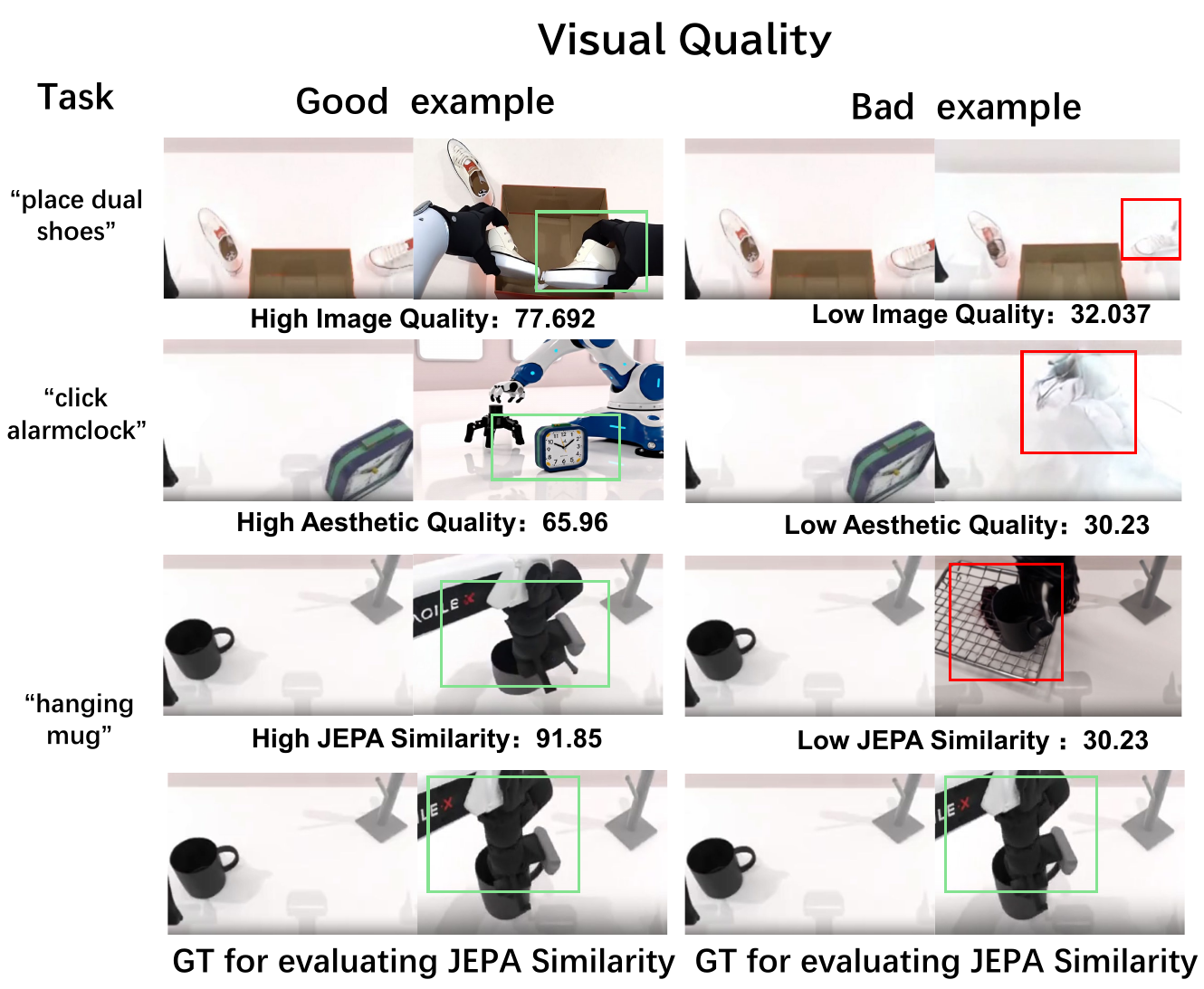} 
    \caption{\textbf{Typical examples of Visual Quality.} \textit{Top:Image Quality.} The bad example on the right-hand-side exhibits significant motion blur and noise, while the good example preserves sharp structural details. \textit{Middle:Aesthetic Quality.} The bad example suffers from severe geometric distortion and artifacts. Conversely, the good example demonstrates superior contrast and realistic lighting with clear reflections. \textit{Bottom:JEPA Similarity.} In the good example, the style and morphology closely align with the GT, while in the bad example, the robotic gripper shows color discrepancies and introduces unintended grid artifacts not present in the GT.
    }
    \label{fig:Vision Quality}
\end{figure*}

\begin{figure*}[h!]
    \centering   \includegraphics[width=0.75\textwidth]{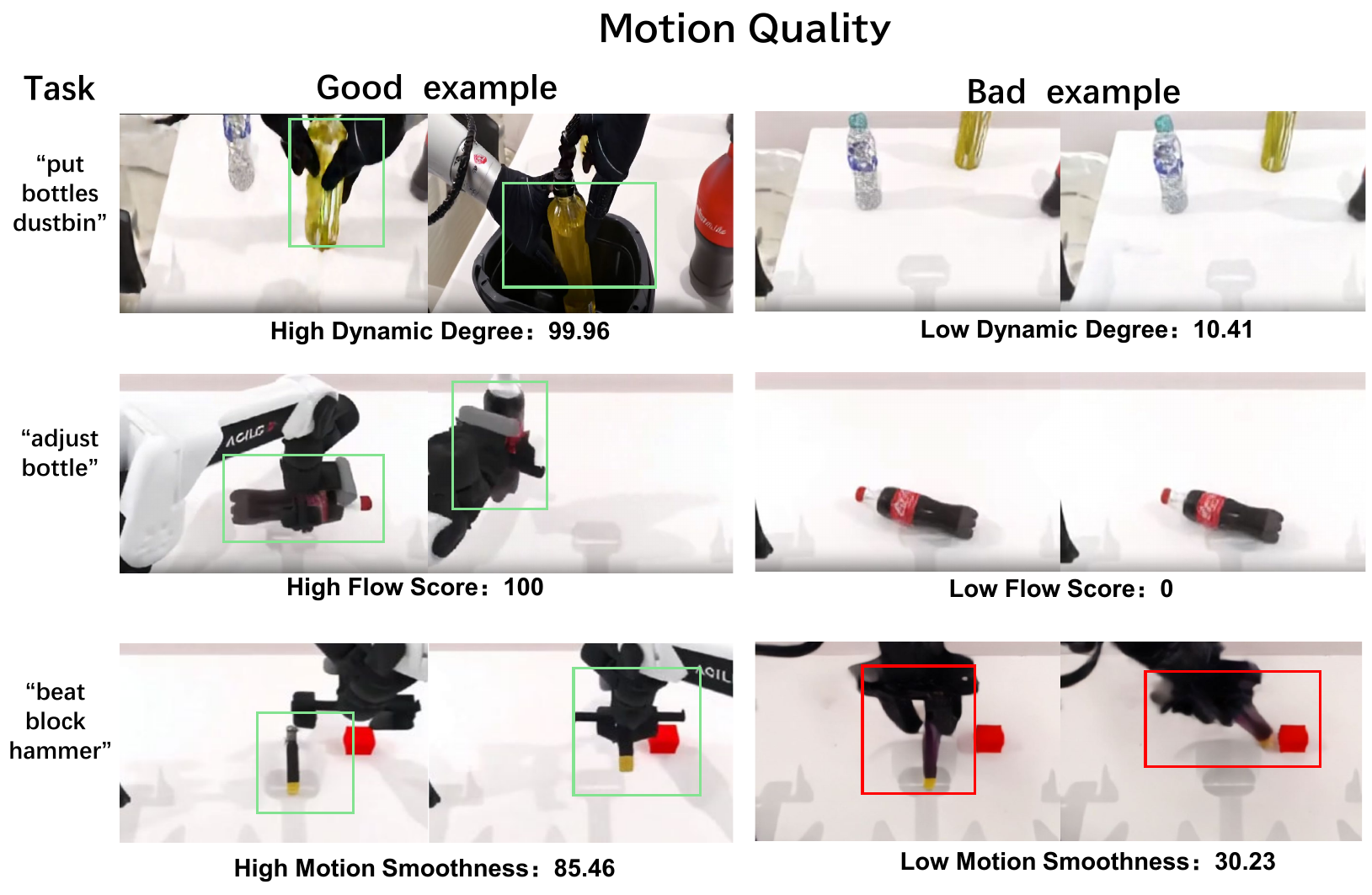} 
    \caption{\textbf{Typical examples of Motion Quality.} \textit{Top:Dynamic Degree.} The good example shows the robotic arm exhibiting a complete and distinct motion sequence from picking up the bottle to placing it in the dustbin, while in the bad example, robotic arm remains static with only minor flickering of the bottle. \textit{Middle:Flow Score.} The good example demonstrates a fluid manipulation of rotating the bottle with significant pixel-level movement and the bad example shows negligible motion, with only slight deformation at the top of the bottle. \textit{Bottom:Motion Smoothness.} The good example features a stable and continuous translation of the hammer, but the bad example suffers from erratic shaking and disjointed,sharp movements immediately after grasping the object.
    }
    \label{fig:Motion Quality}
\end{figure*}
\begin{figure*}[h!]
    \centering   \includegraphics[width=0.75\textwidth]{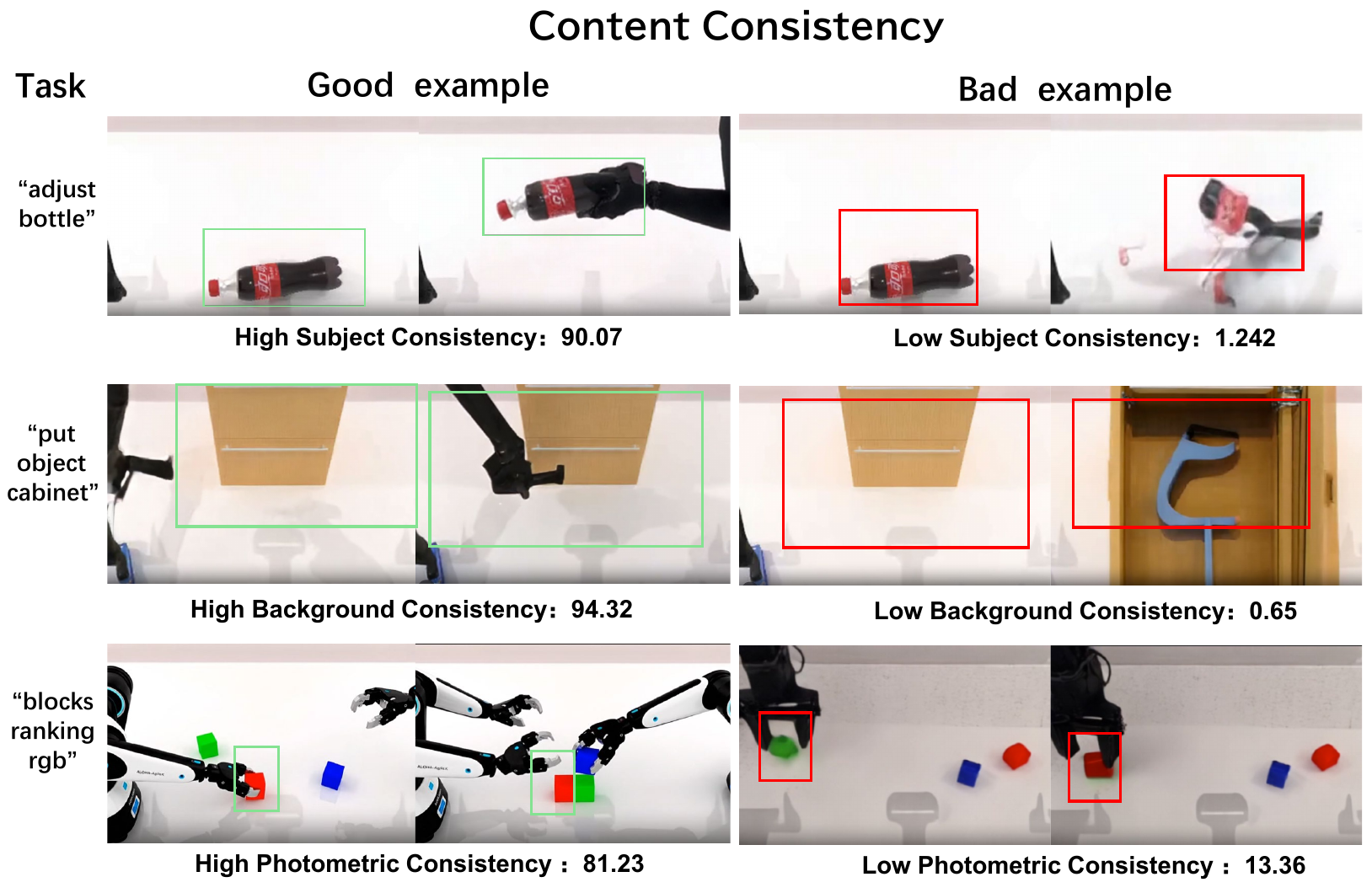} 
    \caption{\textbf{Typical examples of Content Consistency.} \textit{Top:Subjective Consistency.} In the good example, the bottle's shape, color, and packaging remain stable and coherent throughout the grasping process. In the bad example, the bottle suffers from severe deformation and structural chaos, losing its original identity. \textit{Middle:Background Consistency.} The good example maintains a stable background and camera perspective during the cabinet interaction. Conversely, the bad example on the right exhibits a sudden camera shift to a top-down view, leading to an unstable and rapidly changing background. \textit{Bottom:Photometric Consistency.} In the good example, the appearance and color of both the block and the robotic arm are consistently preserved. In the bad example, the grasped block undergoes an unnatural color transition from green to red, indicating poor photometric stability.
    }
    \label{fig:Content Consistency}
\end{figure*}
\begin{figure*}[h!]
    \centering   \includegraphics[width=0.75\textwidth]{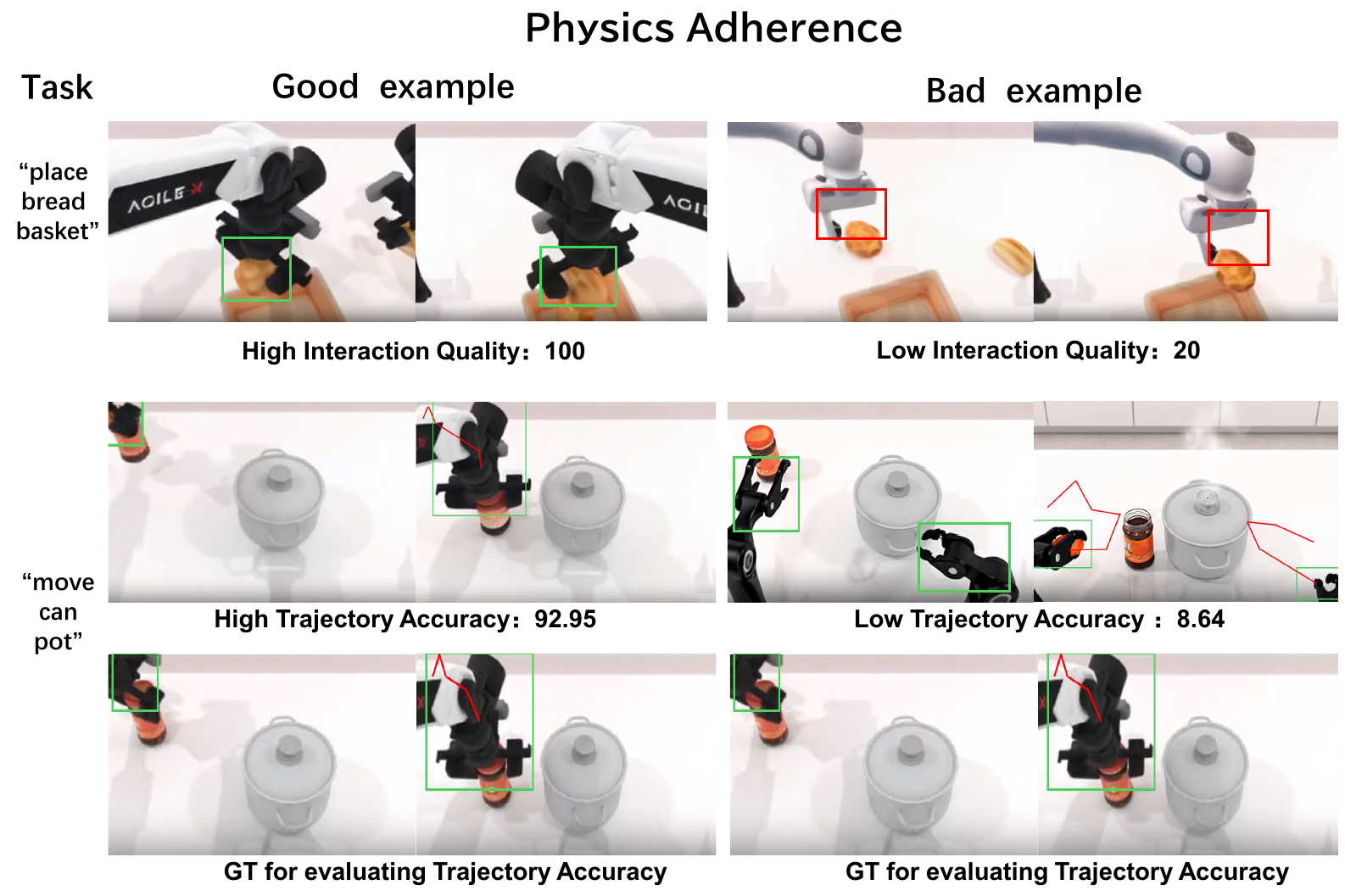} 
    \caption{\textbf{Typical examples of Physics Adherence.} \textit{Top:Interaction Quality.} In the good example, the robotic gripper interacts with the bread appropriately. In the bad example, the bread is lifted without any physical contact with the gripper, violating the fundamental physics laws. \textit{Bottom:Trajectory Accuracy.} The good example demonstrates a movement trajectory that highly aligns with GT. Conversely, the bad example exhibits significant deviations from the GT trajectory, characterized by anomalous movements and jitter.
    }
    \label{fig:Physics Adherence}
\end{figure*}
\begin{figure*}[h!]
    \centering   \includegraphics[width=0.75\textwidth]{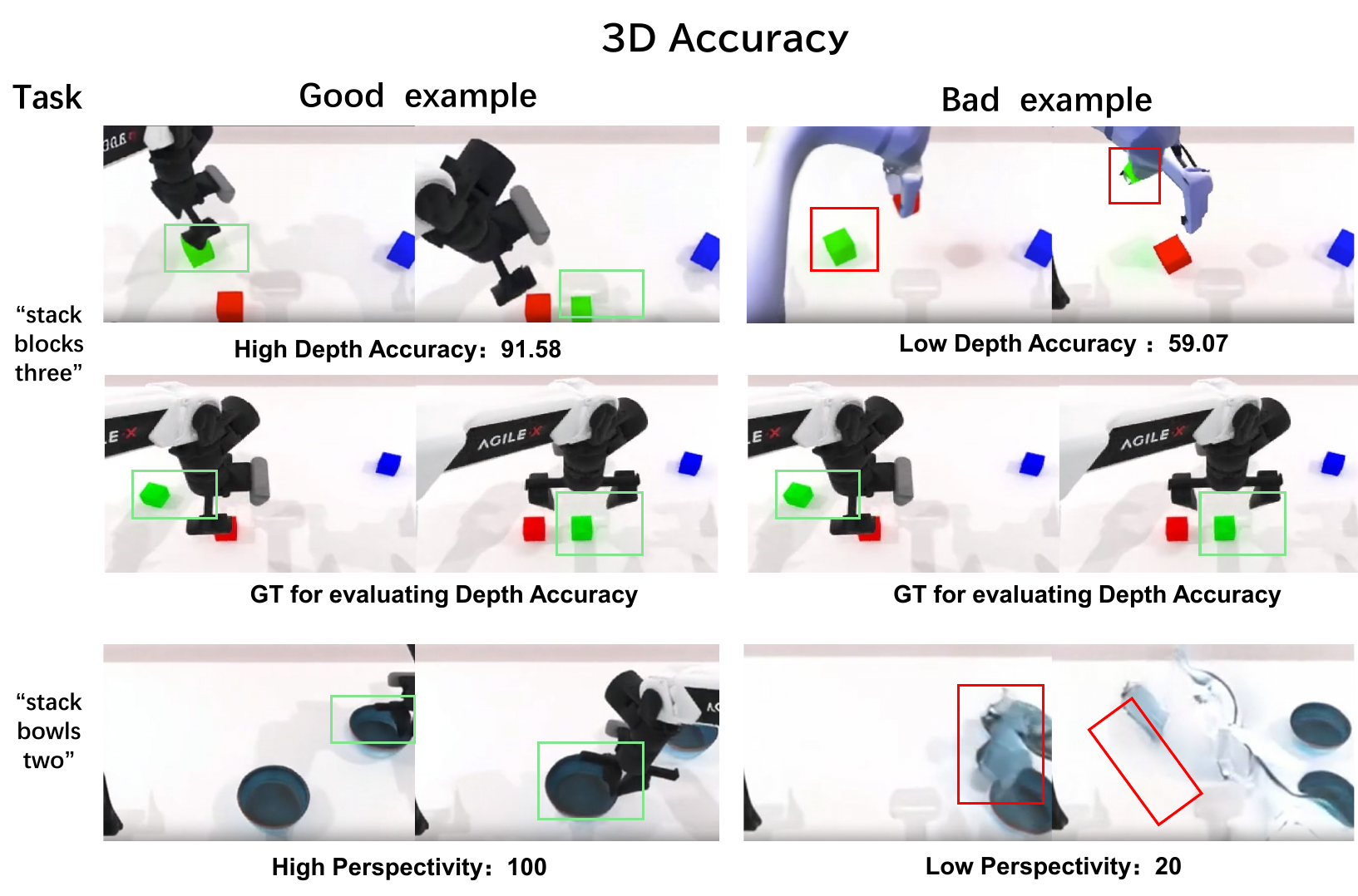} 
    \caption{\textbf{Typical examples of 3D Accuracy.} \textit{Top:Depth Accuracy.} In the good example, the generated depth map highly aligns with the GT, ensuring stable spatial and geometric structures, but the bad example suffers from severe geometric distortion, where the gripper unnaturally merges with the green block, leading to a collapse of spatial integrity. \textit{Bottom:Perspectivity.} The good example maintains realistic perspective and lighting. Conversely, the bad example shows significant ghosting and blurring during movement, failing to preserve the object's contour and exhibiting no shadow of robotic arm that deviate from physical reality.
    }
    \label{fig:3D Accuracy}
\end{figure*}
\begin{figure*}[h!]
    \centering   \includegraphics[width=0.6\textwidth]{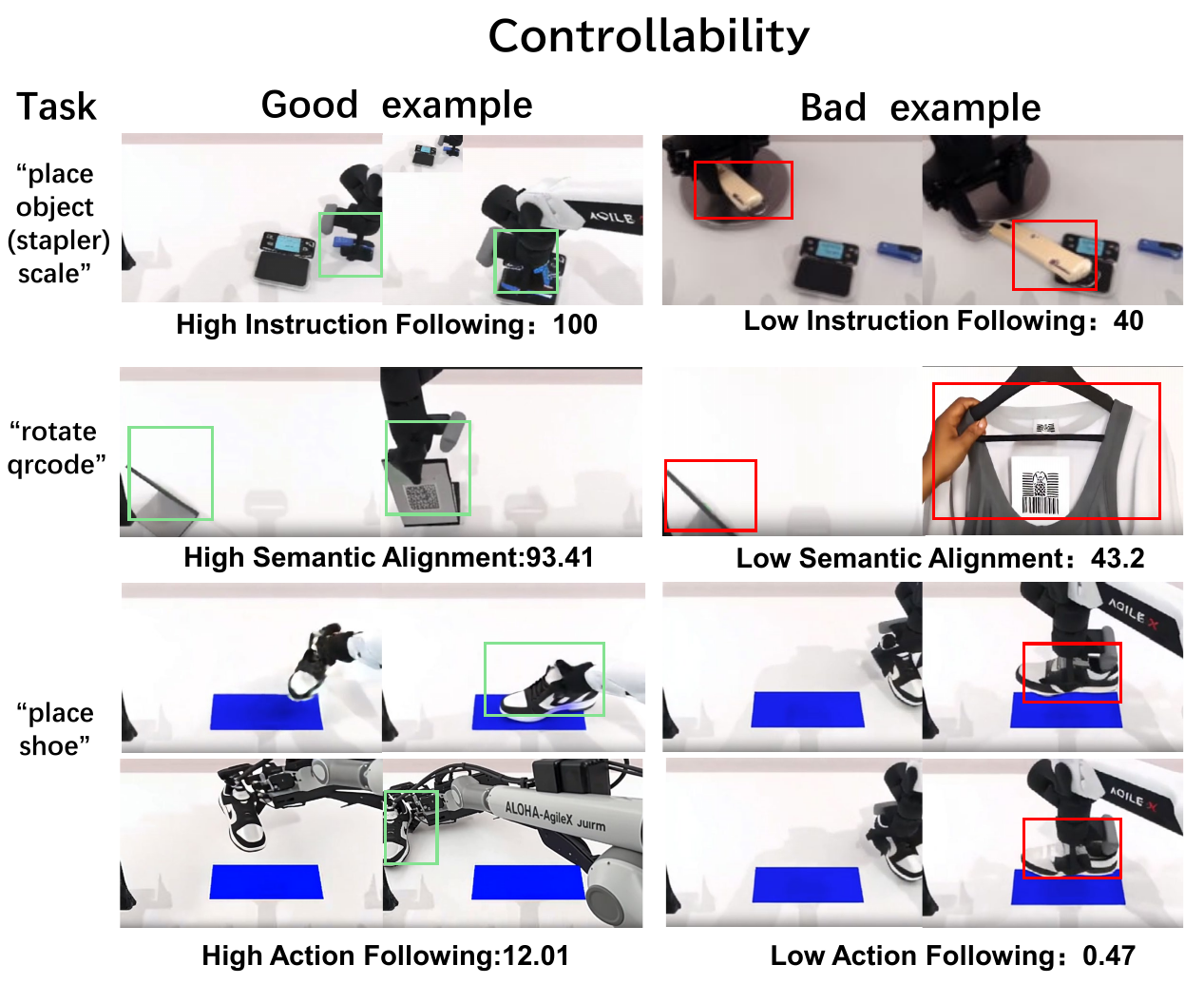} 
    \caption{\textbf{Typical examples of Controllability.} \textit{Top:Instruction Following.} In the good example, the model strictly adheres to the task instruction, but the bad example shows the movement of the incorrect object (knife), failing to execute instruction. \textit{Middle:Semantic Alignment.} The good example demonstrates high semantic fidelity, but the bad example exhibits low alignment by transforming the QR code into a clothing tag and introducing irrational human hands not present in the target semantics. \textit{Bottom:Action Following.} The good example successfully performs distinct actions based on varying prompts, placing the shoe at both the blue marker and to its left, but the bad example shows the model demonstrates limited discriminative ability, executing a singular action regardless of the instruction.}
    \label{fig:Controllability}
\end{figure*}

\newpage

\end{document}